\documentclass[acmtog, nonacm]{acmart}
\usepackage{color}
\usepackage{subcaption}
\usepackage{wrapfig,graphicx}
\usepackage{natbib}
\usepackage{booktabs}
\usepackage{amsmath}
\usepackage[normalem]{ulem}
\usepackage[most]{tcolorbox}
\usepackage{multirow}
\usepackage{makecell}
\usepackage[linesnumbered, ruled, vlined]{algorithm2e}
\usepackage{placeins}

\definecolor{airforceblue}{rgb}{0.36, 0.54, 0.66}
\definecolor{arsenic}{rgb}{0.5, 0.10, 0.8}
\definecolor{darkGreen}{rgb}{0.1, 0.7, 0.1}

\newcommand{\rev}[1]{{#1}}

\begin{document}

\title{Learning Based Toolpath Planner on Diverse Graphs for 3D Printing}

\author{Yuming Huang}\orcid{0000-0001-5900-2164}
\authornotemark[1]
\affiliation{
  \institution{The University of Manchester}
  \city{Manchester}
  \country{United Kingdom}
}

\author{Yuhu Guo}\orcid{0009-0007-5398-6845}
\authornote{Equal contribution of the first two authors.}
\affiliation{
  \institution{The University of Manchester}
  \city{Manchester}
  \country{United Kingdom}
}

\author{Renbo Su}\orcid{0000-0003-0767-6209}
\affiliation{
  \institution{The University of Manchester}
  \city{Manchester}
  \country{United Kingdom}
}

\author{Xingjian Han}\orcid{0000-0002-1899-1952}
\affiliation{
  \institution{Boston University}
  \city{Boston}
  \country{USA}
}

\author{Junhao Ding}\orcid{0000-0001-5741-8115}
\affiliation{
  \institution{The Chinese University of Hong Kong}
  \city{Hong Kong}
  \country{China}
}

\author{Tianyu Zhang}\orcid{0000-0003-0372-0049}
\affiliation{
  \institution{The University of Manchester}
  \city{Manchester}
  \country{United Kingdom}
}

\author{Tao Liu}\orcid{0000-0003-1016-4191}
\affiliation{
  \institution{The University of Manchester}
  \city{Manchester}
  \country{United Kingdom}
}

\author{Weiming Wang}\orcid{0000-0001-6289-0094}
\affiliation{
  \institution{The University of Manchester}
  \city{Manchester}
  \country{United Kingdom}
}

\author{Guoxin Fang}\orcid{0000-0001-8741-3227}
\affiliation{
  \institution{The Chinese University of Hong Kong}
  \city{Hong Kong}
  \country{China}
}

\author{Xu Song}\orcid{0000-0001-9811-9883}
\affiliation{
  \institution{The Chinese University of Hong Kong}
  \city{Hong Kong}
  \country{China}
}

\author{Emily Whiting}\orcid{0000-0001-7997-1675}
\affiliation{
  \institution{Boston University}
  \city{Boston}
  \country{USA}
}

\author{Charlie C. L. Wang}\orcid{0000-0003-4406-8480}
\authornote {Corresponding Author: changling.wang@manchester.ac.uk (Charlie C. L. Wang).  }
\affiliation{
  \institution{The University of Manchester}
  \city{Manchester}
  \country{United Kingdom}
}

\authorsaddresses{
\textit{This is an author's version for research purposes only.}
}

\begin{abstract}
This paper presents a learning based planner for computing optimized 3D printing toolpaths on prescribed graphs, the challenges of which include the varying graph structures on different models and the large scale of nodes \& edges on a graph. We adopt an on-the-fly strategy to tackle these challenges, formulating the planner as a \textit{Deep Q-Network} (DQN) based optimizer to decide the next `best' node to visit. We construct the state spaces by the \textit{Local Search Graph} (LSG) centered at different nodes on a graph, which is encoded by a carefully designed algorithm so that LSGs in similar configurations can be identified to re-use the earlier learned DQN priors for accelerating the computation of toolpath planning. Our method can cover different 3D printing applications by defining their corresponding reward functions. Toolpath planning problems in wire-frame printing, continuous fiber printing, and metallic printing are selected to demonstrate its generality. The performance of our planner has been verified by testing the resultant toolpaths in physical experiments. By using our planner, wire-frame models with up to 4.2k struts can be successfully printed, up to $93.3\%$ of sharp turns on continuous fiber toolpaths can be avoided, and the thermal distortion in metallic printing can be reduced by $24.9\%$.
%
\end{abstract}



\begin{teaserfigure}
\includegraphics[width=\linewidth]{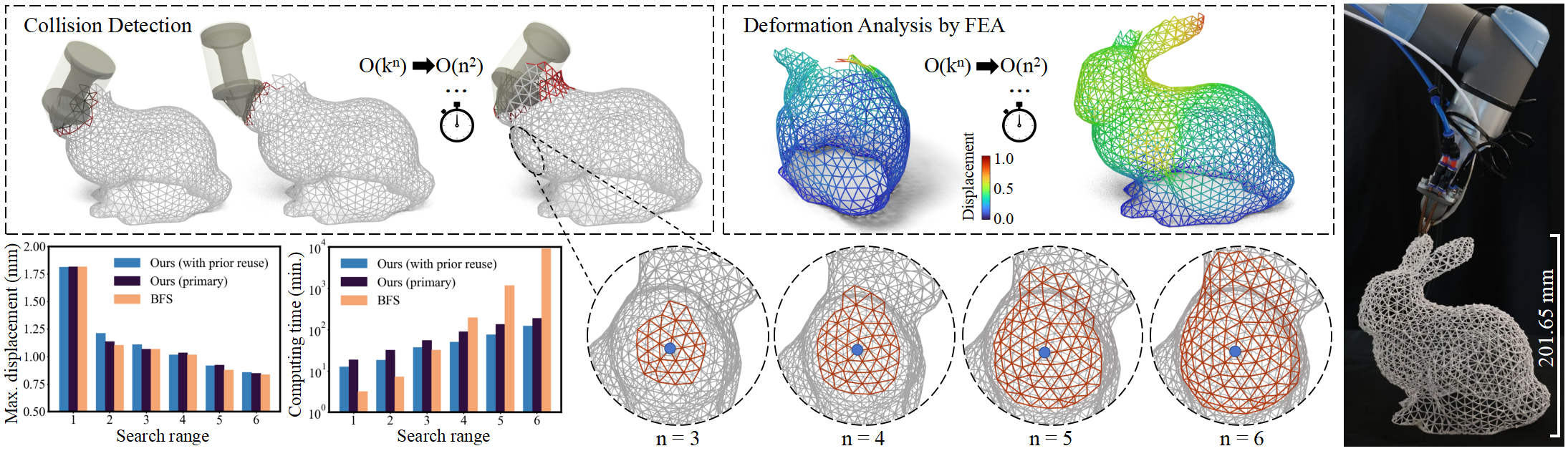}
\centering
\caption{
Our learning based planner suggests an optimized strut sequence with less deformation for wire-frame printing. The planning time is $2.18$ sec. per step (in total $2.05$ hours for the whole model) on a local search graph with $n = 6$ rings of neighbors, which is less than the printing time of around $7.10$ sec. per strut (in total $6.67$ hours). With the same computing time, a \textit{Brute Force Search} (BFS) based planner can only compute on a much smaller local search graph (i.e., $n = 4$ rings) due to its $\mathcal{O}(k^n)$ complexity with $k$ being the valence of a node. Toolpath evaluation involves collision detection and FEA-based deformation analysis. The best toolpath needs to be collision-free and minimizes the maximal displacements on the partially printed structures. Note that the computing times presented in the chart are expressed in logarithmic scale. 
}\label{fig:Teaser}
\end{teaserfigure}

\maketitle

\section{Introduction}\label{secIntroduction}
Toolpath planning problems in 3D printing (3DP) can be formulated as computing an optimized \rev{visiting sequence of nodes} on a given graph, where different models have graph structures with large variation in the number of entities and the topological connectivity. Because of the large diversity and the large scale of graphs to be handled, finding a learning-based solution such as \cite{silver2016mastering,silver2017mastering} to solve this problem on a whole graph is impractical. Differently, this paper proposes a learning-based optimizer by the strategy of `exploration' to work as a `best' next step planner. 

\begin{figure}
\includegraphics[width=\linewidth]{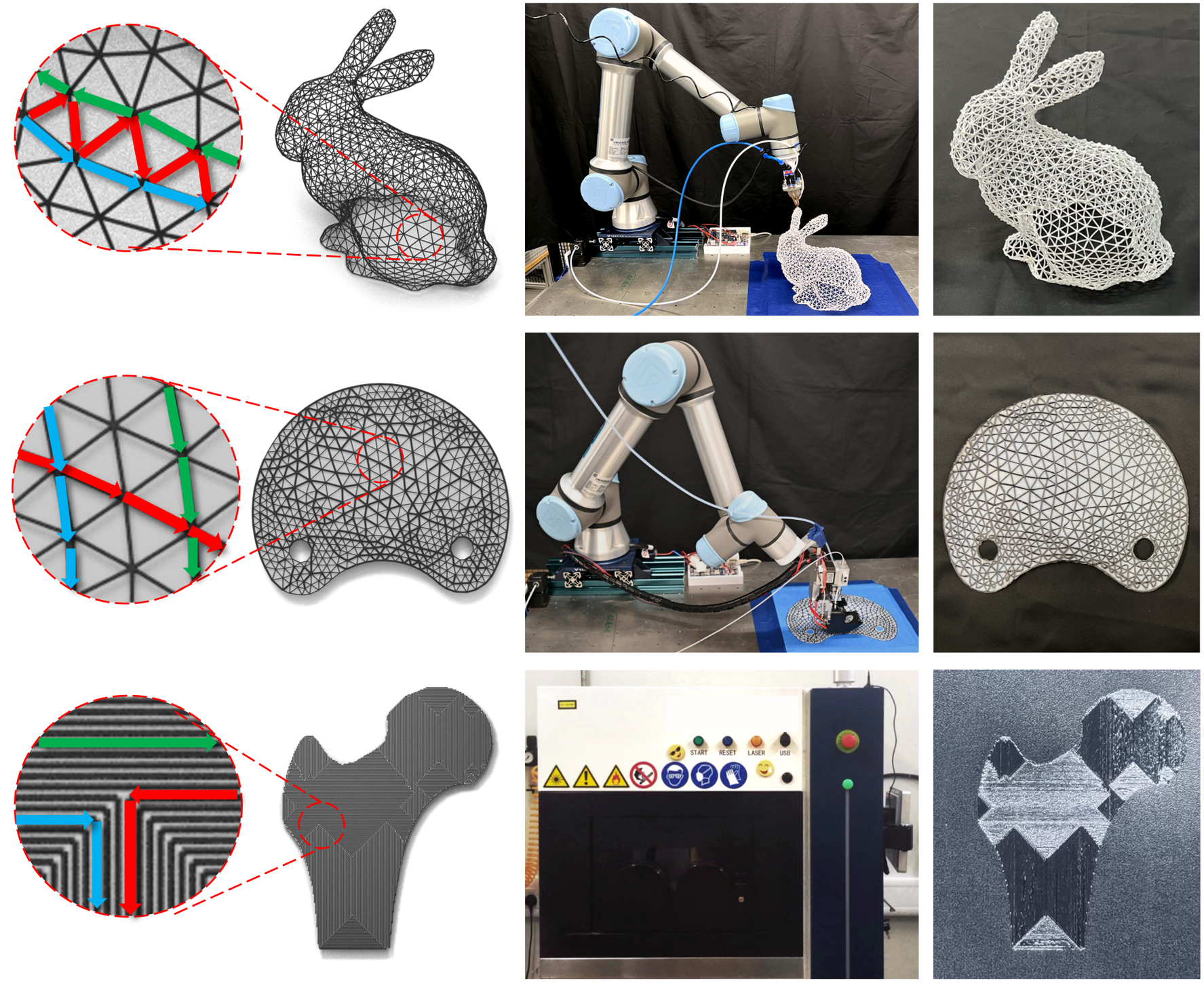}
\put(-243,138){\small \color{black}(a.1)}
\put(-134,138){\small \color{white}(a.2)}
\put(-53,138){\small \color{white}(a.3)}
\put(-243,70){\small \color{black}(b.1)}
\put(-134,70){\small \color{white}(b.2)}
\put(-53,70){\small \color{white}(b.3)}
\put(-243,5){\small \color{black}(c.1)}
\put(-134,5){\small \color{white}(c.2)}
\put(-53,5){\small \color{white}(c.3)}
\vspace{-5pt}
\caption{Applications in computing optimized toolpaths for 3D printing problems of (a) wire-frame models, (b) the continuous fiber reinforced layer for CFRTP, and (c) LPBF-based metal printing. Given input graphs (a.1, b.1 \& c.1), our planner can generate toolpaths optimized according to manufacturing objectives. The toolpaths are tested in physical experiments (a.2, b.2 \& c.2) to produce the results (a.3, b.3 \& c.3). Three different parts of the toolpath are visualized as red, green, and blue arrows.}\label{fig:Three3DPntProblems}
\end{figure}
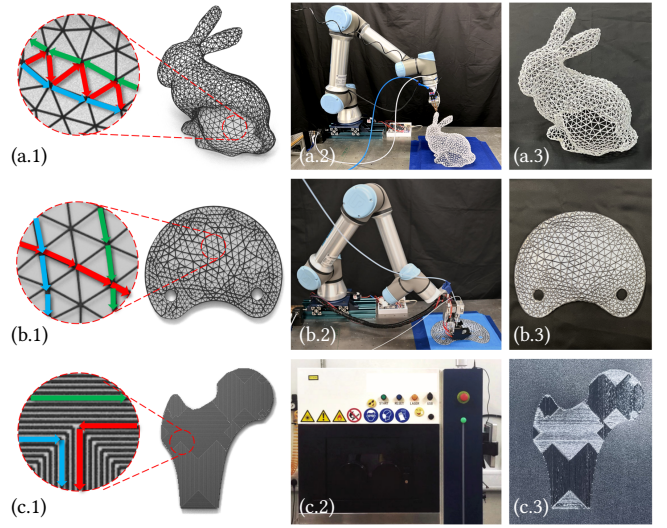

\rev{Many 3DP problems can be formulated as finding an optimized path to visit nodes and edges on} an undirected graph $\mathcal{G}=(\mathcal{E},\mathcal{V})$, which contains a set of edges $\mathcal{E}$ and a set of nodes $\mathcal{V}$. Printing toolpaths are planned to go through all nodes while optimizing different manufacturing objectives. \rev{The resultant toolpath $\mathcal{P}$ is represented as an ordered list of nodes, where their coordinates indicate the tip point of a printer head and their order gives the printing sequence.} Three 3DP problems are considered in this paper, including:  
\begin{enumerate}
\item Wire-frame structures with deformation control and collision avoidance, \rev{giving $\mathcal{G}$ as the target wire-frame structure to be fabricated where every \rev{edge} in $\mathcal{E}$ have to and can only be printed once along the straight trajectory defined by the positions of each edge's two ending nodes (see Fig.\ref{fig:Teaser})}; 

\item \textit{Continuous carbon fibers} (CCF) in \textit{continuous fiber reinforced thermoplastic} (CFRTP) with sharp-turn prevention, \rev{where $\mathcal{G}$ represents the structure (i.e., a 2D wire-frame model) to be formed by CCF filaments between layers of other thermoplastic materials (named as matrix materials)}\footnote{Note that we focus on a manufacturing problem about finding a path to realize the planned structure more reliably here rather than designing a CCF structure with better mechanical strength (e.g.,~\cite{jiang2014freeform,wang2020globally}).};

\item \textit{Laser powder bed fusion} (LPBF) based metallic printing with reduced thermal warpage, \rev{where every planar layer is first rasterized into a binary image 
and the pixels of which are employed as the nodes $\mathcal{V}$ with the edges $\mathcal{E}$ of the graph $\mathcal{G}$ being defined between every nodes in $\mathcal{V}$ to its four neighbors (i.e., left, right, up \& down)}. 
\end{enumerate}
Different 3DP problems will raise different coverage requirements for edges and nodes. We propose a \rev{learning}-based planner that selects nodes from $\mathcal{V}$ to add into the toolpath $\mathcal{P}$ one by one -- see Fig.\ref{fig:Three3DPntProblems} for the example toolpaths for these 3DP problems. 

To solve the planning problem on a large-scale graph, we construct the on-the-fly \textit{Local Search Graph} (LSG) and progressively determine the best choice in an LSG, \rev{where each LSG is a local structure consisting of $n$ rings of nodes and edges around a current node (see the regions with different rings highlighted in orange color in Fig.\ref{fig:Teaser})}. The gap between a local optimum and the global optimum can be narrowed down when using a larger LSG. However, the feasible search range of an LSG is constrained by the allowed maximal time for determining the best next step in an LSG. The range can be very small when the move is determined by the \textit{Brute Force Search} (BFS) that involves time-consuming evaluations such as collision detection and FEA-based deformation analysis for wire-frame 3D printing \cite{huang2016framefab}. Existing approaches are mainly based on heuristics such as greedy selection \cite{wu2016printing} or depth-first search with backtracking~\cite{huang2016framefab,huang2023turning}. The results are in general less optimal than BFS. 

We argue that BFS in an LSG can be replaced by a learning-based planner that provides decisions with similar quality but at much higher efficiency. Specifically, each LSG is converted into a state space employed as an input of a \textit{Deep Q-Network} (DQN) based planner. The output of this planner is the best next step determined according to the reward function. Progressively applying this planner from node to node, an optimized toolpath can be generated on a graph, the scale of which can be very large (e.g., with up to 10k nodes in our experiments) due to the on-the-fly nature of the planner. Given an LSG with $n$-rings of nodes around the current center, the computational complexity is reduced from $\mathcal{O}(k^n)$ to \rev{$\mathcal{O}(n^2)$} when changing from BFS to a DQN based planner with $k$ being the valence of a node. Here, we assume the node valences on a given graph have a small variation. An algorithm is developed to encode nodes in LSGs so that similar state spaces can be formed when LSGs have similar configurations. This enables the re-use of priors (i.e., neural networks) trained in earlier DQN based learning, which accelerates the computation of our on-the-fly planner. As shown in Fig.~\ref{fig:Teaser} for an example of wire-frame 3D printing, this allows to use an LSG with $n=6$ while still achieving a real-time online planner to determine the best next-strut -- i.e., the planning time (around $2.05$ hours in total) is much shorter than the printing time (around $6.67$ hours for the whole model).

The technical contributions of our work are as follows:
\begin{itemize}
\item We propose an efficient DQN based on-the-fly planner (see Sec.~\ref{subsecToolpathGen}) for computing optimized 3D printing toolpaths, which is feasible for diverse graphs in large scale.

\item We investigate an algorithm to construct the state spaces that preserve the state similarity for LSGs with similar configurations (Sec.~\ref{subsecStateSpace}), which enables the prior reuse (Sec.~\ref{subsecPriorReuse}).

\item Different reward functions are defined for the manufacturing objectives of different 3D printing processes (Sec.~\ref{secReward}). 
\end{itemize}
Physical experiments are conducted on a variety of models to demonstrate the performance of our method in different applications. 

\section{Related work}\label{secRelatedWork}

\begin{figure*}[t]
\centering
\includegraphics[width=\linewidth]{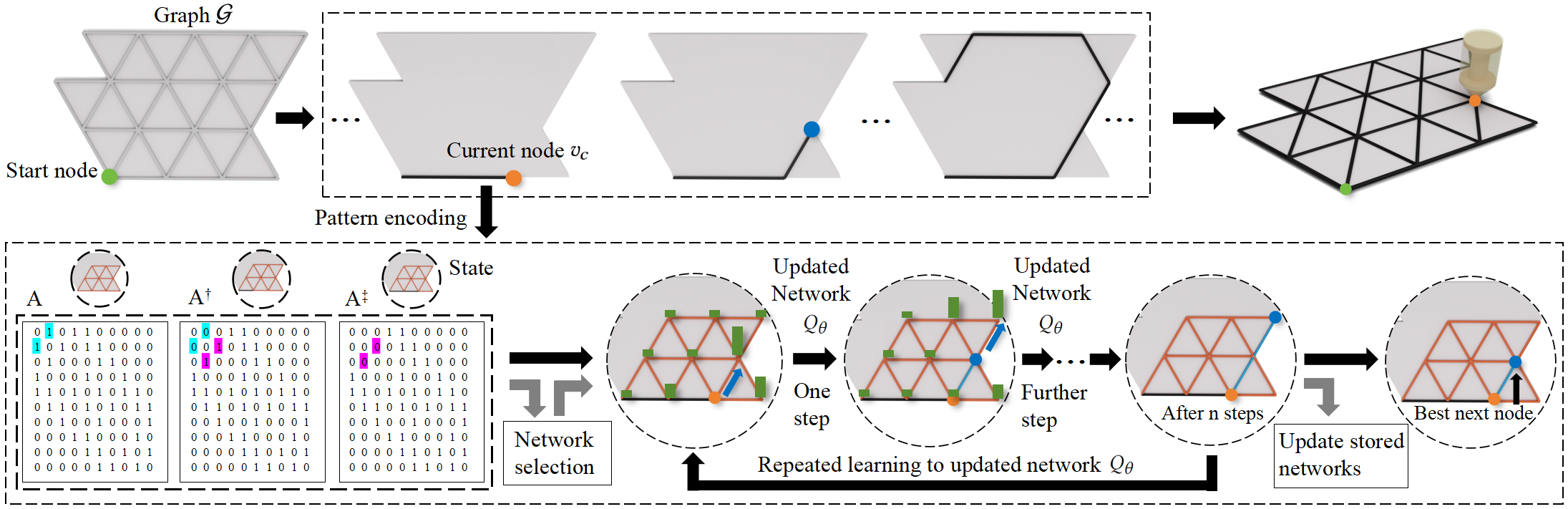}
\put(-506,152){\small \color{black}(a)}
\put(-400,152){\small \color{black}(b)}
\put(-312,152){\small \color{black}(c)}
\put(-222,152){\small \color{black}(d)}
\put(-110,152){\small \color{black}(e)}
\put(-506,78){\small \color{black}(f)}
\put(-320,78){\small \color{black}(g)}
\put(-60,78){\small \color{black}(h)}
\caption{(a)-(e) Illustration of using our $Q$-learning based planner to compute the toolpath on a graph $\mathcal{G}$. Our on-the-fly planner explores $\mathcal{G}$ on a LSG with $n$-rings of neighbors ($n=2$ in this example), given a current node $v_c$ (orange in (b,g,h)). (f) The moving state $\mathbf{S}$ of $v_c$ is a 3D matrix commonly determined by the current LSG and two previous steps in the partially planned toolpath (in black), which is formed by three adjacency matrices $\mathbf{A}$, $\mathbf{A}^{\dag}$ and $\mathbf{A}^{\ddag}$. (g, h) The next best node of $v_c$ is determined by learning an updated deep $Q$-network to compute the $Q$-values (green bars) on every nodes in the LSG according to the state $\mathbf{S}^*$. Color blocks in (f) highlight the differences between $\mathbf{A}$ and $\mathbf{A}^{\dag}$ (pink color) and between $\mathbf{A}^{\dag}$ and $\mathbf{A}^{\ddag}$ (blue color). The best next node from (h) will update to become next current node in (c). Repeatedly applying this $Q$-learning based planner can determine the resultant toolpath as shown in (e).
}\label{fig:AlgOverview}
\end{figure*}
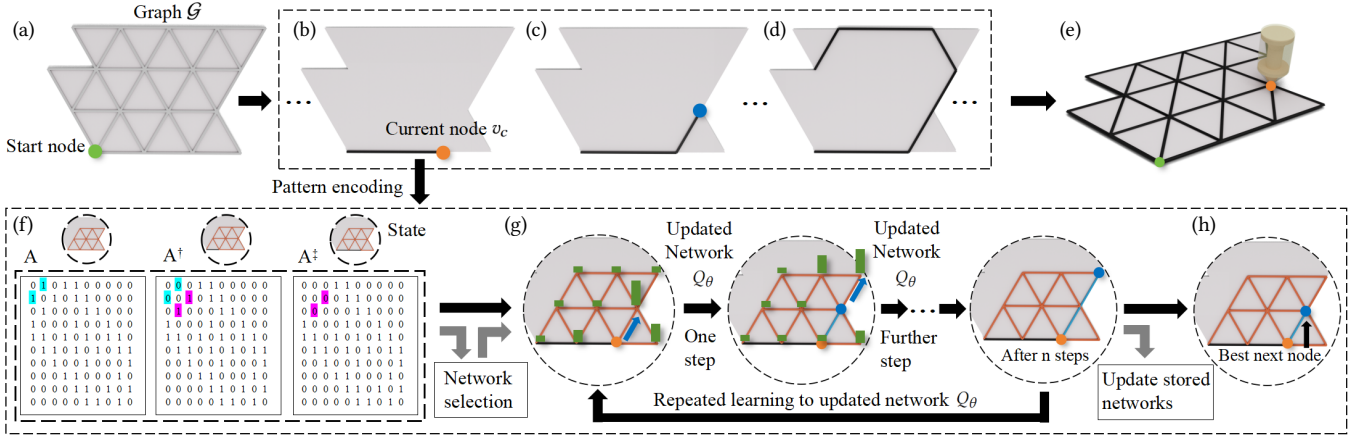

\subsection{Toolpath generation for manufacturing}
Toolpath generation, as a critical step to enable the 3D printing process, has caught a lot of attention in the community of computational fabrication. The toolpaths are often generated to completely fill a given 2D region while preserving the continuity and being conformal to the boundary of the 2D region. \citet{zhao2016connected} presented a method to generate continuous toolpaths from the boundary distance field as contour-parallel Fermat spirals, which satisfy both requirements. The concept is later extended to compute milling toolpaths for 3D surfaces \cite{zhao2018dscarver}. A recent work \cite{bi2022continuous} can generate continuous contour-zigzag hybrid toolpaths with both the solid and partial infill patterns. Field based methods have been developed to generate toolpaths that can well use the anisotropic mechanical strength of filaments (ref.~\cite{fang2020reinforced,chen2022field}). Differently, we will focus on toolpath planning instead of toolpath generation in this paper.

We can find a variety of travel planning approaches in the literature that optimize different aspects of toolpaths, including collision \cite{wu2016printing}, turning angles~\cite{zhang2023graph}, fiber consumption \cite{sun2023more}, energy-efficiency \cite{pavanaskar2015energy} and thermal distribution \cite{ramani2022smartscan}. Toolpath planning algorithm relies heavily on the domain-specific knowledge of  materials being processed and the objective functions to be optimized. A general framework of toolpath planning that can effectively handle different objectives is not available yet.

\subsection{Heuristic methods}
Planning an optimized path on a given graph can be an NP-hard problem in many cases (e.g.,~\cite{gupta2021sfcdecomp}). There is no guarantee that an optimal solution can be obtained in polynomial time \cite{hochba1997approximation}. To obtain the planning results on large scale graphs, heuristics are applied to compute the `local' optimum. 

For the problem of CCF toolpath generation, \citet{yamamoto2022novel} proposed a method generating continuous, single-stroke paths using Eulerian paths and Hierholzer’s algorithm. \citet{zhang2022graph} incorporated the measurement against large turning angles by integrating local constraints into Fleury's algorithm. The length and the quality (i.e., the number of sharp turns) of toolpaths generated by these methods are not well optimized. To reduce the thermal distortion in metallic 3D printing, an adaptive greedy toolpath generation method was introduced in \cite{qin2023adaptive} based on the heuristic of moving melting point away from the recently-processed regions. Again, the resultant toolpaths are only optimized locally.

To avoid time-consuming BFS algorithms that can provide global optimum on a graph (e.g.,~\cite{gao2019near}), researchers have conducted the heuristics of using \textit{depth first search} (DFS) algorithms with backtracking. These algorithms can achieve better results than greedy heuristics in general. Example approaches include the work of robot-assisted wire-frame printing \cite{huang2016framefab} and the CCF toolpath planning \cite{huang2023turning}. However, these DFS-based algorithms still do not work on graphs in large scales due to the exponential growth of the potential paths.

\subsection{Learning-based methods}
Learning-based methods provide new opportunities to solve the path planning problem. They can be classified into two major categories, supervised learning and \textit{reinforcement learning} (RL). Supervised learning approaches (e.g., \cite{kim2022tool,nguyen2020continuous}) always need a large dataset to complete the learning process. For example, a point completion network is employed in \cite{Wu2019DeepNBVPlanner} to build a deep-learning based next-best view planner for robot-assisted 3D reconstruction. However, in the 3DP application with graphs in diverse structures, acquiring a sufficiently large dataset for supervised learning is very challenging. 

Differently, RL-based methods transform the path planning problem into a Markov decision process. This strategy has been widely employed in the area of robotics for solving the problems of target-drive mapless navigation \cite{pfeiffer2018reinforced} and path planning \cite{lv2019path}. Researchers (e.g.,~\cite{kool2018attention,joshi2022learning}) have employed RL-based methods to solve the famous \textit{traveling salesman problem} (TSP), which is a typical path planning issue. Previous research shows that the RL-based methods have an upper bound of the optimality gap better than the supervised learning \cite{silver2017mastering,joshi2022learning}. They are also much less expensive than \textit{Brute Force Search} (BFS) in computing time when the number of nodes in the graph is larger than 100. While these approaches have successfully addressed the problem of dataset-free and whole-graph global optimization, it is still a challenging task to apply them to graphs on large scales (e.g., those with more than 10k nodes considered in our work). Specifically, they are applied to small graphs with less than 200 nodes where the training already needs more than 400 hours. The dynamic window strategy needs to be employed together to make learning-based methods scalable. 

\subsection{Planning with dynamic windows} 
The strategy of planning with dynamic windows has been widely employed in the path planning of robotic research to explore unknown environments. Rapidly exploring random tree (RRT) algorithm is a typical example that updates the path in an incremental manner based on sampling \cite{lavalle2001randomized}. RRT's computational cost can be very expensive if the local search range is large. To this end, \citet{chiang2019rl} proposed an RL-based local obstacle avoiding planner, which is based on a supervised-learning based reachability estimator. The concept of dynamic windows has been employed in other path planning approaches (e.g., \cite{chang2021reinforcement,wang2020mobile,ogren2005convergent}). However, 

\rev{they focused on finding a feasible path to reach a target position, which is different from 3DP toolpath planning problems for obtaining an optimized path to cover a given graph.}

RL-based methods are also employed to realize the tasks of online control and compensation, which can also be considered as dynamic window based (i.e., in terms of time window). For example, it is employed to control the motion of soft manipulators in \cite{thuruthel2018model}. \rev{RL-based method has been employed to solve the graph search problem in \cite{Zhao2020} and determine optimal policies of multi-material printing in \cite{Liao2023}.} 
Recently, RL-based approach has been applied in \cite{piovarvci2022closed} to correct the printing path in real time by using images as feedback. Differently, we develop a $Q$-learning based planner to compute optimized toolpaths based on the moving states defined on LSGs, which leads to a general framework that can handle a variety of 3D printing applications by defining different reward functions.

\section{Learning based planner}\label{secPlanner}
\subsection{Preliminary: DQN-based reinforcement learning}\label{subsecDQNRL}

\rev{The toolpath planning problem can be modeled as a Markov decision process with state space, action space, transition function and reward function, which can be efficiently computed by integrating Q-learning with a deep neural network (ref.~\cite{silver2016mastering,silver2017mastering}). Specifically, a network parameterized by its network coefficients $\theta$ is learned to approximate the $Q$-value function as $Q_{\theta}(\mathbf{S},\mathbf{a})$ that takes the state $\mathbf{S}$ as input and outputs $Q$-values for any possible action $\mathbf{a}$. These $Q$-values represent the expected cumulative rewards for different actions. This is inspired by prior research in reinforcement learning \cite{mnih2015human, mnih2013playing, lillicrap2015continuous} so that the resultant network $Q_\theta$ can help predict the `best' next action. In our formulation, the key components of the DQN-based learning process are defined as follows.}
\begin{itemize}
\item \rev{\textit{State Space}: The state space is parameterized as 3D matrices representing the configuration of nodes in a LSG and the short-term memory of configurations from the previous two steps. Further details are discussed in Sec.~\ref{subsecStateSpace}.}

\item \rev{\textit{Action Space}: The action space consists of possible moves from the center node of a LSG to all other nodes within the same LSG.}

\item \rev{\textit{Transition function}: The transition function defines how the current state changes to the next state simulating the printing process. This is done using FEA and collision detection for wire-frame printing, estimating sharp-turning angles for CCF printing, and computing temperature distributions for metallic printing. More details are provided in Sec.~\ref{subsecManuObjectives}.}

\item \rev{\textit{Reward function}: The reward function evaluates how well the manufacturing objectives are achieved. Detailed formulas for this function are provided in Sec.~\ref{subsecRewardFormulation}.}
\end{itemize}
\rev{With a well-trained network $Q_{\theta}$, selecting the action that maximizes the function value will also maximize the cumulative reward.}

\rev{We now introduce the learning process to obtain an optimized $Q_{\theta}$. After giving an initialized network, the training starts from an initial state $\mathbf{S}_0$. Taking a random action $\mathbf{a}$ on a state $\mathbf{S}$ will give a new state $\mathbf{S}'$, which will also lead to a reward $r$. Different applications of reinforcement learning learn can have different reward functions as that we define in Sec.~\ref{subsecRewardFormulation}. This forms a sample of the experience $(\mathbf{S},\mathbf{a},R,\mathbf{S}')$.
The samples are employed to train the neural network $Q_{\theta}(\cdot)$ for estimating Q-values that satisfy the Bellman equation as}
\begin{equation}\label{eqBellman}
    y = r + \gamma \max_{\mathbf{a}^*} \left\{ \bar{Q}(\mathbf{S}',\mathbf{a}^*) \right\},
\end{equation}
\rev{which calculates the target Q-value $y$ with $\gamma \in (0,1)$ being the discount factor. The network $Q_{\theta}(\cdot)$ is optimized by performing gradient descent steps to minimize the loss as}
\begin{equation}\label{eqRLLoss}
    L = (y - Q(\mathbf{S},\mathbf{a}))^2.
\end{equation}
\rev{The network's weights $\theta$ are updated to minimize this loss, improving the $Q$-value estimates. Note that in the Bellman equation, the $Q$-network $\bar{Q}_{\theta}(\cdot)$ is called \textit{target} network that is only updated by $Q_{\theta}(\cdot)$ at regular intervals (e.g., every 10 steps) to address the challenge of instability and divergence in training. The sophisticated DQN-learning schemes also include the steps of storing samples in a replay buffer and sampling random mini-batches of experiences from the replay buffer to train the neural network.  
}

\begin{algorithm}[t]
\SetAlgoLined 
\caption{\rev{DQNBasedToolpathPlanner}}
\KwIn{An undirected graph $\mathcal{G}=(\mathcal{E},\mathcal{V})$}
\KwOut{The toolpath $\mathcal{P}$ as a list of nodes to be visited}

Select a starting node $v_c$ and let $\mathcal{P} \leftarrow \{v_c\}$;

\Repeat{the graph $\mathcal{G}$ has been fully explored}{

Construct the LSG $\mathcal{G}_c$ centered at $v_c$;

Establish the state $\mathbf{S}$ of $v_c$ by $\mathcal{G}_c$ and $\mathcal{P}$ (Sec.~\ref{subsecStateSpace});

Prior selection to initialize the network $Q_\theta$ (Sec.~\ref{subsecPriorReuse});

\tcc{\rev{Optimize $Q_\theta$ by exploring different next $n$ moves (DQN-based reinforcement learning)}}

\Repeat{the terminal condition of learning is satisfied}{

Compute the $Q$-values of all nodes in $\mathcal{G}_c$ by the current network $Q_\theta$ and the state $\mathbf{S}$;

Assign a state $\mathbf{S}^1 \leftarrow \mathbf{S}$;

\While{$i=1,2,\cdots,n$}{

\rev{Choose the next node $v^i_c$ by $\epsilon$-greedy strategy;}

\rev{Compute the reward value $r_i$ of an action $\mathbf{a}_i$ if adding $v_c^i$ into $\mathcal{P}$;}

\rev{Compute a new state $\mathbf{S}^{i+1}$ by extending $\mathcal{P}$ to $v^i_c$;}
}

\rev{Optimize $Q_\theta$ by Eqs.(\eqref{eqBellman} \& \eqref{eqRLLoss}) using $n$ samples obtained as $(\mathbf{S}^i,r_i,\mathbf{a}_i,\mathbf{S}^{i+1})$ ($i=1,2,\cdots,n$);}

}

Update the prior as a set of stored networks (Sec.~\ref{subsecPriorReuse});

Compute the $Q$-values of all nodes in $\mathcal{G}_c$ by $Q_\theta$;

Among the one-ring neighbors of $v_c$, select the best next node $v_c^*$ according to the largest $Q$-value;

Add $v_c^*$ into $\mathcal{P}$ and let $v_c \leftarrow v_c^*$;
}

\Return $\mathcal{P}$;

\end{algorithm}

\subsection{Toolpath generation algorithm}\label{subsecToolpathGen}
We now present how to compute an optimized toolpath by an on-the-fly $Q$-learning based planner. Our toolpath generation algorithm randomly selects a node in $\mathcal{V}$ as the first node, assigned as $v_c$, for planning the toolpath. After constructing the state $\mathbf{S}$ of $v_c$ by the method \rev{presented in Sec.~\ref{subsecStateSpace}}, a $Q$-learning algorithm is applied to the LSG $\mathcal{G}_c$ centered at $v_c$ to generate an optimized \textit{deep $Q$-learning network} (DQN). The $Q$-value of all nodes in $\mathcal{G}_c$ are then computed by the optimized DQN, among which the best next node is selected among $v_c$'s 1-ring neighbors. Repeatedly applying this Q-learning based planner determines the resultant toolpath. 

The \rev{pseudo-code} of our algorithm is given in \rev{Algorithm \textit{DQNBasedToolpathPlanner}} (see also the illustration given in Fig.~\ref{fig:AlgOverview}). For applications with special requirements, the starting node of our algorithm can be selected from a subset of the input graph -- e.g., a toolpath for wire-frame printing can only start from the bottom of the structure. The terminal condition of toolpath generation also depends on different 3D printing applications. Different graph coverage requirements will be presented in Sec.~\ref{subsecManuObjectives}. \rev{The terminal condition for the interior routine of DQN-based reinforcement learning (i.e., Lines 6-14 in Algorithm \textit{DQNBasedToolpathPlanner}) will be discussed in Sec.~\ref{subsecTerminalCond}.}

\subsection{Moving state representation}\label{subsecStateSpace}
\rev{Our toolpath planner suggests} the best next node to visit on $\mathcal{G}$ according to the state $\mathbf{S}$ defined on an LSG. Starting from a \rev{current} node $v_c \in \mathcal{V}$, a breadth-first search is conducted to find $n$-rings of neighbor nodes around $v_c$ and store them in a set $\mathcal{V}_c$. We can then construct a set of edges $\mathcal{E}_c$, where every edge in $\mathcal{E}_c$ should have its two nodes in $\mathcal{V}_c$. The LSG centered at $v_c$ is then defined as $\mathcal{G}_c=(\mathcal{E}_c,\mathcal{V}_c)$. Given an LSG $\mathcal{G}_c$ with $m$ nodes, we can convert it into an adjacency matrix $\mathbf{A}= [a_{i,j} ]_{m\times m}$ where the value of $a_{i,j}$ defines the length of edge connecting the nodes $v_i, v_j \in \mathcal{V}_c$ divided by the maximal edge length in the same LSG to normalize its value. And $a_{i,j}=0$ is given if 1) there is no edge between $v_i$ and $v_j$, or 2) the edge has been included in the toolpath $\mathcal{P}$. In other words, $a_{i,j}=0$ means that the toolpath is not allowed to include the travel between $v_i$ and $v_j$. \rev{An example matrix $\mathbf{A}$ can be found in Fig.\ref{fig:AlgOverview}(f).} 

\begin{figure}
\includegraphics[width=\linewidth]{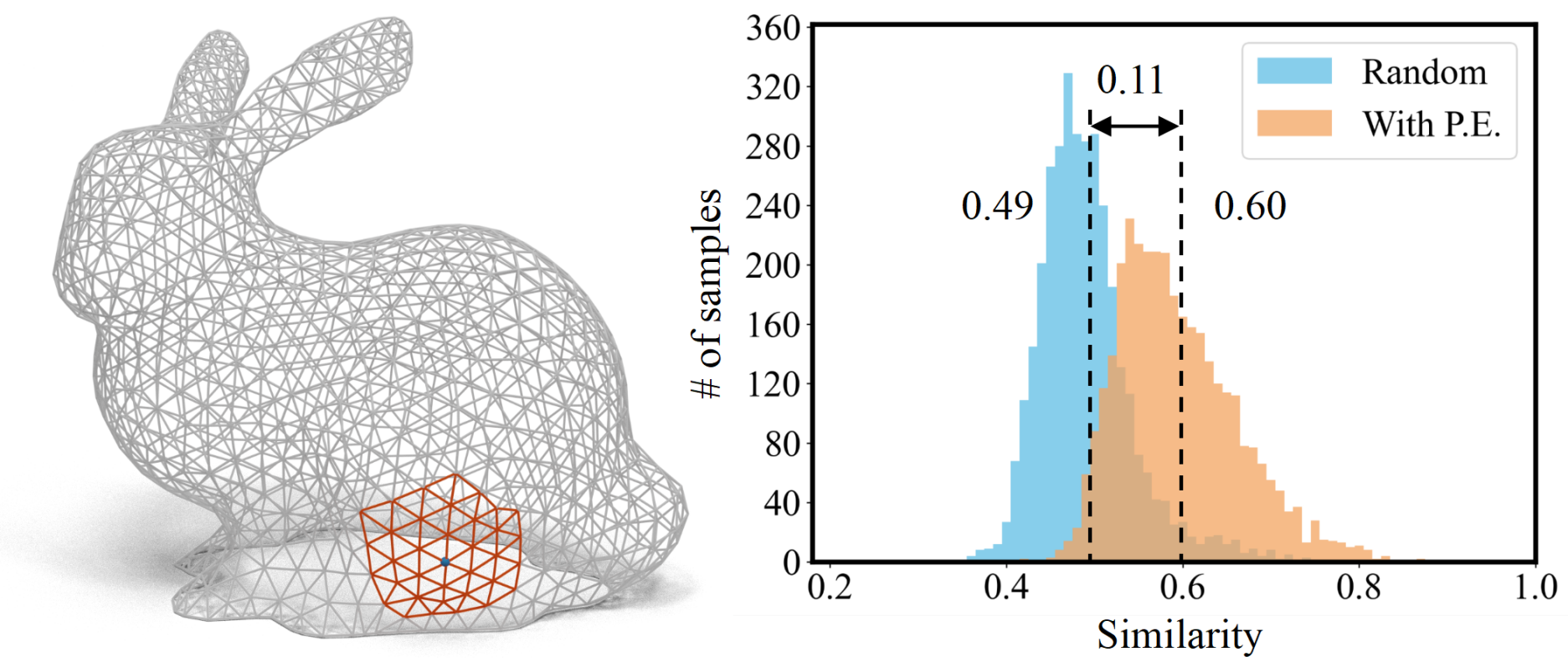}
\caption{A study to verify the effectiveness of the pattern encoding method. Given a LSG as shown on the Bunny model, we compare the similarity of its adjacent matrix to the other adjacent matrices formed when visiting other nodes by a random order. The histograms of similarities with vs. without the pattern encoding (P.E.) algorithm are compared.
}\label{fig:patternStudyMatrix}
\end{figure}

\subsubsection{Pattern encoding} 
The pattern of an adjacent matrix generated by aforementioned method heavily rely on the indices of nodes given in $\mathcal{G}_c$. We introduce a simple yet effective algorithm to index the nodes and therefore similar adjacent matrices can be obtained for LSGs with similar configurations. \rev{To determine the order of nodes in a LSG,} each LSG is first flattened into a planer graph with the positions of $v_q$ and $v_c$ being fixed at $(-1,0)$ and $(0,0)$ respectively. The order of nodes are then sorted and indexed by their $x$- and $y$-coordinates on the \rev{flattened mesh}. The conformal parameterization method \cite{Levy2002LSCM} that can preserve the local shape conformity is employed in our implementation. As found from the study shown in Fig.\ref{fig:patternStudyMatrix}, the similarity of adjacent matrices -- measured by the metric defined in Eq.(\ref{eqSimilarityMetric}) can be significantly improved with this indexing method, which consistently encodes the similarity of LSGs into the pattern similarity of adjacent matrices. Note that fixing both $v_q$ and $v_c$ in parameterization will prevent the unwanted rotation when encoding the pattern of LSG configurations into adjacent matrices. More experimental tests for verifying the effectiveness of the pattern encoding algorithm can be found in the supplementary document.

\subsubsection{Short-term memory information}\label{subsecShortMemory}
The matrix $\mathbf{A}$ provides a basic state space of the LSG and the already formed path $\mathcal{P}$. Without loss of generality, we assume that the last three nodes of $\mathcal{P}$ are $v_p$, $v_q$ and $v_c$ in $\mathcal{G}_c$ \rev{with the order of $v_p \rightarrow v_q \rightarrow v_c$}. 
To encode the short-term memory information of $\mathcal{P}$ into the state space, we extended $\mathbf{A}$ into a 3D matrix as $\mathbf{S}=[\mathbf{A}, \mathbf{A}^{\dag}, \mathbf{A}^{\ddag}]_{m \times m \times 3}$ with $\mathbf{A}^{\dag}$ and $\mathbf{A}^{\ddag}$ being the adjacency matrices of the previous step and two steps ago. Specifically, $\mathbf{A}^{\dag}$ is a copy of $\mathbf{A}$ but adding the normalized length $l_{q,c}$ of the previous step $v_q \rightarrow v_c$ into the corresponding elements as $a_{q,c}=a_{c,q}=l_{q,c}$. Similarly, $\mathbf{A}^{\ddag}$ can be obtained from $\mathbf{A}^{\dag}$ by adding the normalized length $l_{p,q}$ of the earlier step $v_p \rightarrow v_q$ as $a_{p,q}=a_{q,p}=l_{p,q}$. This strategy is very common in DQN-based learning, especially in problems with temporal dependencies where the current action depends on a sequence of previous states rather than just the current state (ref.~\cite{janner2021offline}). When a partially determined $\mathcal{P}$ only has two nodes or one node, we define $\mathbf{S}$ as $[\mathbf{A}, \mathbf{A}^{\dag}, \mathbf{A}^{\dag}]$ and $[\mathbf{A}, \mathbf{A}, \mathbf{A}]$ respectively. 

During the iterative learning process introduced in the following sub-section, we will explore the rewards of different actions in the LSG and thus to update the deep $Q$-network for $Q$-value prediction. When the action is to add a new edge $v_r \rightarrow v_s$, where $v_r, v_s\in \mathcal{V}_c$, the state $\mathbf{S}$ will be changed from $[\mathbf{A}, \mathbf{A}^{\dag}, \mathbf{A}^{\ddag}]$ to $[\mathbf{A}^*, \mathbf{A}, \mathbf{A}^{\dag}]$ with $\mathbf{A}^*$ being obtained from $\mathbf{A}$ by making $a_{r,s}=a_{s,r}=0$. This indicates a new state that the edge has been part of $\mathcal{P}$. \rev{The effectiveness of this representation for the state with short-term memory information has been verified by comparing with the results only using $\mathbf{A}$ as the state -- see details provided in the supplementary document. Both efficiency and performance can be improved by incorporating historical information into the state representation.}

For an LSG with regular node valences (e.g., valence $k=6$ for the graph as shown in Fig.~\ref{fig:AlgOverview}), the number of nodes in an LSG \rev{is bounded by $1+3n(n+1)$ -- that is $\mathcal{O}(n^2)$.}
According to experiments, we find that $m=50n$ safely works for all examples in our tests. This is mainly because the node valences on example graphs tested in our experiments are often less than 7. 

\subsection{Acceleration scheme of toolpath planner}\label{subsecPriorReuse}
The $Q$-learning algorithm can always generate an optimized DQN after exploring enough possible paths (ref.~\cite{silver2017mastering}). Whether the learning process can converge quickly depends on the initial coefficient $\theta$ employed for $Q_\theta$. \rev{Without using the steps of prior selection and update (i.e., Lines 5 \& 16 in Algorithm \textit{DQNBasedToolpathPlanner})}, the initial value of $\theta$ is based on the learning result of the previous node (i.e., $v_q \in \mathcal{P}$), the graph configuration of which can be significantly different from the current node $v_c$. Therefore, the learning process on many nodes resembles starting a new search each time. We introduce an accelerated scheme to allow DQN reuse, which is benefited by the method encoding the graph patterns into the states as the distributed values in adjacent matrices. \rev{Note that our acceleration scheme is based on network reuse, which is different from the sample-based experience relay in conventional DQN-based learning (ref.~\cite{silver2016mastering}).}

Without loss of the generality, we can assume that $K$ different \rev{networks} have been stored as $\Theta=\{\theta_k\}$ together with their corresponding states $\mathcal{S}=\{\mathbf{S}_k\}$ with $k=1,\ldots,K$. After obtaining the state $\mathbf{S}$ for the LSG centered at $v_c$ \rev{(i.e., Line 4 of Algorithm \textit{DQNBasedToolpathPlanner})}, we compare the similarity between $\mathbf{S}$ and $\{\mathbf{S}_k\}$ by
\begin{equation}\label{eqSimilarityMetric}
    \rho_k = \rho(\mathbf{S},\mathbf{S}_k)  = \frac{1}{\lambda} \left(1 + \| \mathbf{S} - \mathbf{S}_k\|_F \right)^{-1}
\end{equation}
with $\| \cdot \|_F$ being the Frobenius norm. $\lambda$ is a coefficient that can be employed to control the scale of $\rho_k$ \rev{with $\lambda=0.76$ being used in all our tests}. Among all these $K$ stored \rev{networks}, the one with its configuration most similar to $\mathbf{S}$ is selected as the initial value of $\theta$ to learn an updated DQN. That is to assign $\theta \equiv \theta_s$ with $s = \arg \max_k \{ \rho_k\}$. The $Q$-learning 
\rev{by reusing previously learned networks with} 
similar configuration can effectively reduce the number of iterations for reaching the termination criterion. \rev{This has been verified by the ablation study given in the supplementary document.} 

The toolpath planner starts from no prior. The first $K$ nodes will have their states and the optimized network coefficients added into $\mathcal{S}$ and $\Theta$. After that, the newly learned \rev{network} needs merge into the set of \rev{stored networks} by the following method.
\begin{itemize}
\item First of all, the most similar pair of states denoted by $\mathbf{S}_a$ and $\mathbf{S}_b$ are selected among these $(K+1)$ states;

\item For $\mathbf{S}_a$ and $\mathbf{S}_b$, their maximal similarities to the other $(K-1)$ states are evaluated as $\rho_a$ and $\rho_b$; 

\item When $\rho_a > \rho_b$, remove $\mathbf{S}_a$ and $\theta_a$ from the set while keeping the other $K$ \rev{networks}; otherwise, remove $\mathbf{S}_b$ and $\theta_b$.
\end{itemize}
This selection strategy will allow us to keep good diversity of states among the stored \rev{networks}. While choosing a small $K$ reduces the effectiveness of acceleration (i.e., \rev{networks} from dissimilar states will be used), a large number $K$ will also bring in the cost of comparing time and memory consumption. We choose $K=10$ in all examples to achieve the balance between effectiveness and cost. 

\section{Reward functions}\label{secReward}
Our learning-based toolpath planner is general and can be applied to a variety of 3D printing problems. This section first analyzes the manufacturing objectives of three different 3D printing processes, including wire-frame, CFRTP, and LPBF-based metal. After that, their corresponding reward functions for learning are introduced.

\subsection{Manufacturing objectives}\label{subsecManuObjectives}

\subsubsection{Wire-Frame Models}~To realize the 3D printing of wire-frame models by a robotic-arm with 6 degree-of-freeform (DOF) motions, the following manufacturing objectives need to meet on the resultant toolpath $\mathcal{P}$.
\begin{itemize}
\item \textit{Coverage:} Each edge of the input graph must be passed exactly once;

\item \textit{Collision:} There is a collision-free solution for the robotic-arm's trajectory when realizing the material deposition for each edge;

\item \textit{Deformation:} The structural deformation (caused by gravity) should be minimized on the finally and all partially completed structures. 
\end{itemize}
First of all, the continuity of the toolpath is not required for printing wire-frame models. Instead, we need to avoid printing the same edge more than once as this will lead to poor printing quality. Moreover, at any stage of the fabrication process, the printer head and the robotic arm should not collide with the partial structure that has been printed or the environmental obstacles (ref.~\cite{wu2016printing}). 

Deformation minimization is very critical for printing large wire-frame models as the displacement on nodes caused by gravity can be accumulated. When a large displacement occurs, the newly printed struts cannot be connected to the already printed partial structure that has been deformed away from its planned positions (see Fig.~\ref{fig:failureWireframePeeling} for an example). Specifically, when using a printer head with nozzle dimension as $1.0\mathrm{mm}$, the maximal displacement on a partially printed structure should be less than $1.0\mathrm{mm}$. Here the deformation of a structure needs to be evaluated repeatedly during the toolpath planning process by FEA. 

\begin{figure}
\centering
\includegraphics[width=\linewidth]{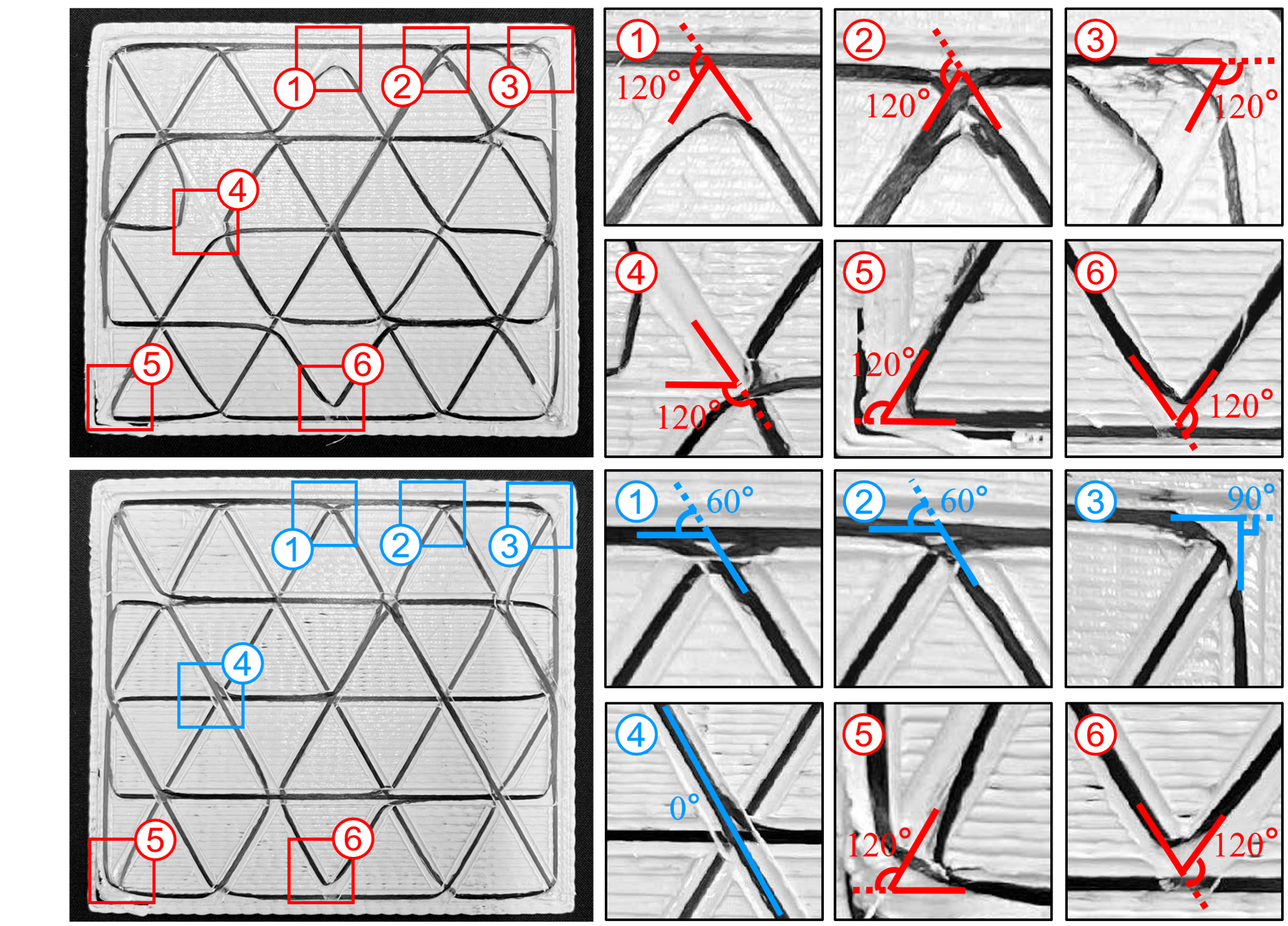}
\put(-242,91){\small \color{black}(a)}
\put(-242,3){\small \color{black}(b)}
\caption{Printing CCF along an unoptimized toolpath generated by (a) DFS on the input graph leads to 6 sharp turns (regions in red squares -- failure of carbon fiber adhesion), which can be reduced into (b) 2 sharp turns by using our planner. The turning angles in blue squares have been changed so that improved fiber adhesion can be observed. 
}\label{fig:failureCCFSharpTurn}
\end{figure}

\vspace{5pt}
\subsubsection{Continuous Fibers for CFRTP}\label{subsubsecCCF}~3D printing techniques have been employed to fabricate composites, where CCF are used to make reinforcement layers between thermal plastic layers. To reduce the usage of CCF while still providing strong reinforcement, the reinforcement layer is in the form of structures that has strong mechanical strength (e.g., the triangular / honeycomb cells \cite{jiang2014freeform} or the grids following the streamlines of principal stresses \cite{wang2020globally}). The manufacturing objectives of CCF toolpaths in the reinforcement layers are as follows.
\begin{itemize}
\item \textit{Coverage}: Each edge of the input graph must be fabricated by one to two passes of CCF deposition while minimizing total material usage;

\item \textit{Continuity}: All passes are continuously connected -- i.e., the whole graph can be covered by CCF in `one-stroke';

\item \textit{Sharp-turn}: The CCF toolpath should \rev{avoid} sharp turns.
\end{itemize}
The first objective is to fabricate the structure completely while minimizing the usage of CCF. Discontinuity on the toolpath of CCF can significantly reduce the mechanical strength of CFRTPs thus needs to be prevented. However, only Euler graphs can satisfy the topological requirement to find a complete travel path with continuity \cite{leiserson1994introduction}. We then relax the condition to allow an edge being traversed twice, where this relaxation enlarges the search space to determine a toolpath with fewer sharp turns.

Printing CCF along toolpaths with curvatures larger than the bundle size can result in defects (see Fig.~\ref{fig:failureCCFSharpTurn}) due to the strong axial stiffness of fiber materials \cite{matsuzaki2018effects}. We need to prevent sharp turning angles exceeding $120^{\circ}$, which can lead to misalignment, breakage, folding, and thickness inconsistencies \cite{zhang2021fibre,sanei20203d}. All lead to weaker mechanical strength on 3D printed CFRTPs. Moreover, turning angles between $60^{\circ}$ and $120^{\circ}$ need to be reduced when possible. 

\vspace{5pt}
\subsubsection{LPBF-based Metal Printing}
\begin{figure}
\centering
\includegraphics[width=\linewidth]{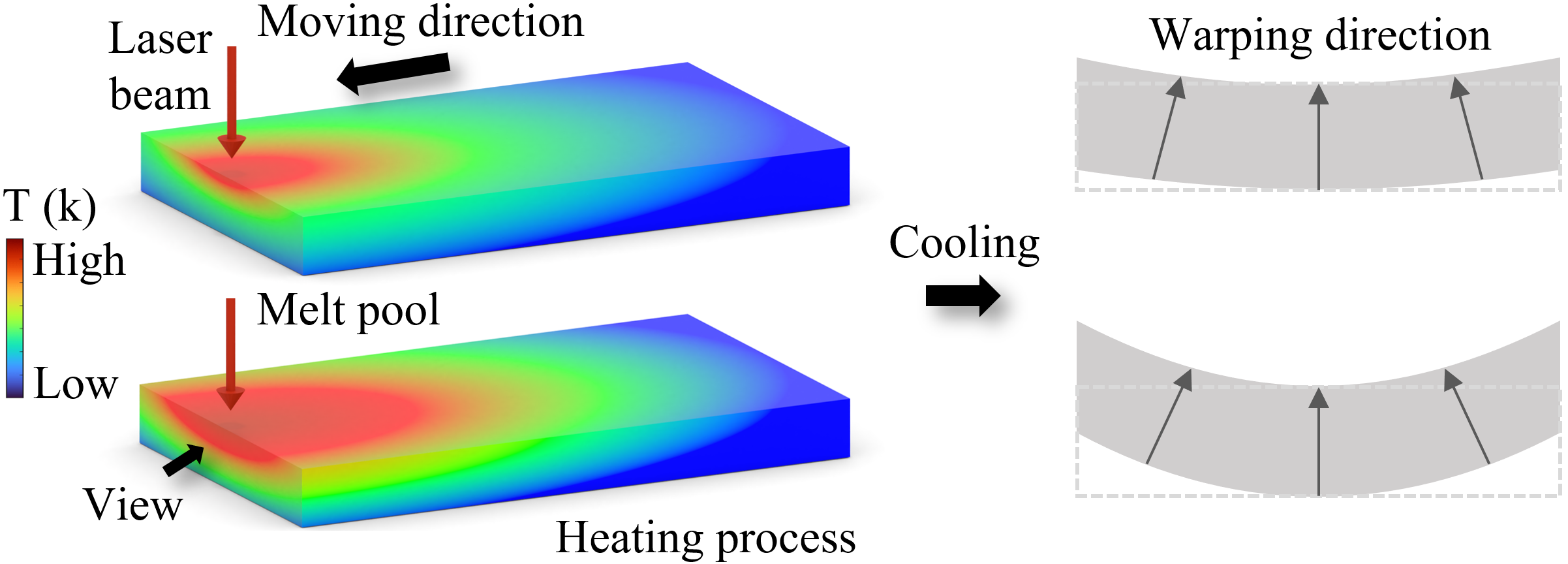}
\caption{The illustration of the warpage caused by concentrated thermal stresses in the metal printing process. During the cooling process, materials in large melt pools need to undergo larger temperature gradients and corresponding volume changes, which is the main cause of warpage.}
\label{fig:wrap}
\end{figure}

\begin{figure}
\centering
\includegraphics[width=\linewidth]{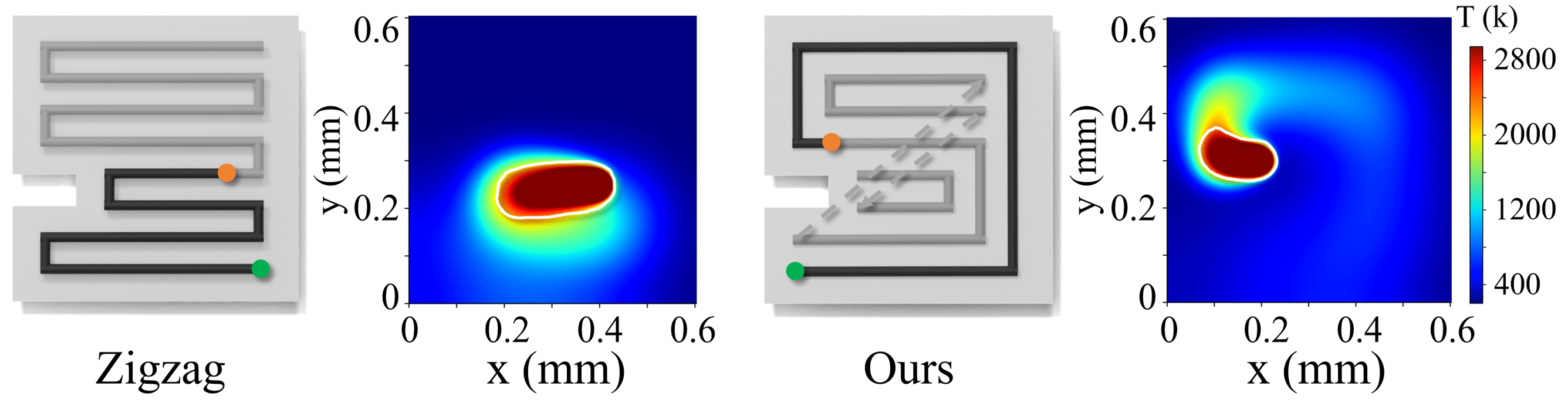}
\caption{An illustration of different temperature distributions of the powder bed after moving the melting laser beam for the same duration but along different toolpaths from the green point to the orange point: (left) zigzag toolpath and (right) a toolpath generated by our planner. The regions with high temperature are enclosed by the isocurves $T = 2600$k. Our toolpath gives a region with $43 \%$ smaller area.
}\label{fig:metal_simu}
\end{figure}

~A problem of LPBF-based metallic printing is the deformation caused by thermal stresses \cite{ramani2022smartscan}. Specifically, when the toolpath (i.e., the source of laser melting) enters a region with high temperature, a larger melt pool will be generated which needs to undergo larger temperature gradients and corresponding volume changes. In short, this leads to large thermal stress locally and is the major source of thermal warpage (see Fig.~\ref{fig:wrap}). During the powder fusion process, a larger region with high temperature leads to a larger melt pool. Different patterns of toolpaths will give different temperature distributions, which has been demonstrated by a simple example shown in Fig.~\ref{fig:metal_simu}. 

The manufacturing objectives for LPBF-based metallic printing include coverage and thermal uniformity as explained below.
\begin{itemize}
\item \textit{Coverage}: Every node of the graph must be covered exactly once;

\item \textit{Thermal Uniformity}: A temperature field will be generated by laser melting when driving the laser beam along a toolpath. The region of high temperature needs to be minimized.
\end{itemize}
During the process of LPBF-based metallic printing, visiting the same point multiple times will lead to heat accumulation and should be avoided. Therefore, a node on the given graph is only allowed to be visited once. The thermal uniformity is also demanded to avoid heat accumulation, where the temperature field can be evaluated by a numerical simulation \cite{qin2023adaptive} that is simplified by only incorporating the decay and diffusion of heat on printed nodes. 

\subsection{Formulations}\label{subsecRewardFormulation}
\begin{wrapfigure}[7]{r}{0.28\linewidth}\vspace{-20pt}
\begin{center}
\hspace{-25pt}\includegraphics[width=1.0\linewidth]{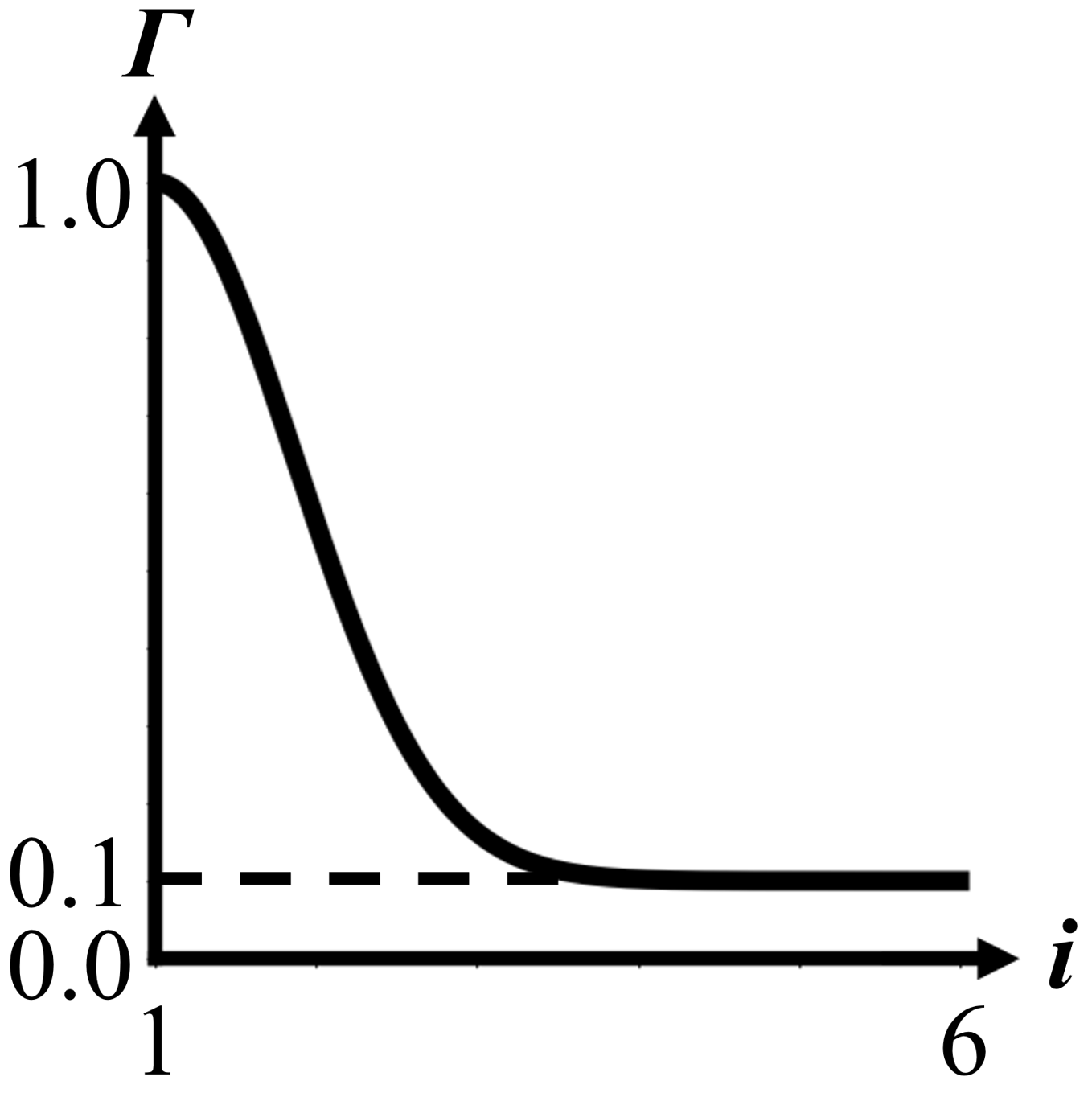}
\end{center}
\end{wrapfigure}
We now present the formulations of reward functions for different 3D printing applications. Besides of the manufacturing objectives discussed in the above sub-section, the discount factor of different steps on an LSG is also considered in our planner. Specifically, the moving action closer to the center $v_c$ of an LSG will have a larger weight than the steps away -- i.e., the closer the more important. For the $i$-th step ($i=1,2,\cdots,n$), the discount function is defined as
\begin{equation}\label{eq:DecayFunc} 
    \Gamma(i)= 0.9 \exp \left( -\frac{(i-\mu)^2}{2\sigma^2} \right) + 0.1
\end{equation}
with $\sigma$ and $\mu$ being parameters of Gaussian. $\sigma=0.865$ and $\mu=1.0$ are chosen by \rev{experimentation} with the function curve shown above. \rev{Note that this choice of discount function is different from conventional DQN-based reinforcement learning \cite{silver2016mastering,silver2017mastering}. A comparison for verifying the effectiveness of our `Gaussian'-like discount function can be found in the supplementary document.}

\begin{wrapfigure}[10]{r}{0.4\linewidth}\vspace{-10pt}
\begin{center}
\hspace{-10pt}\includegraphics[width=1.0\linewidth]{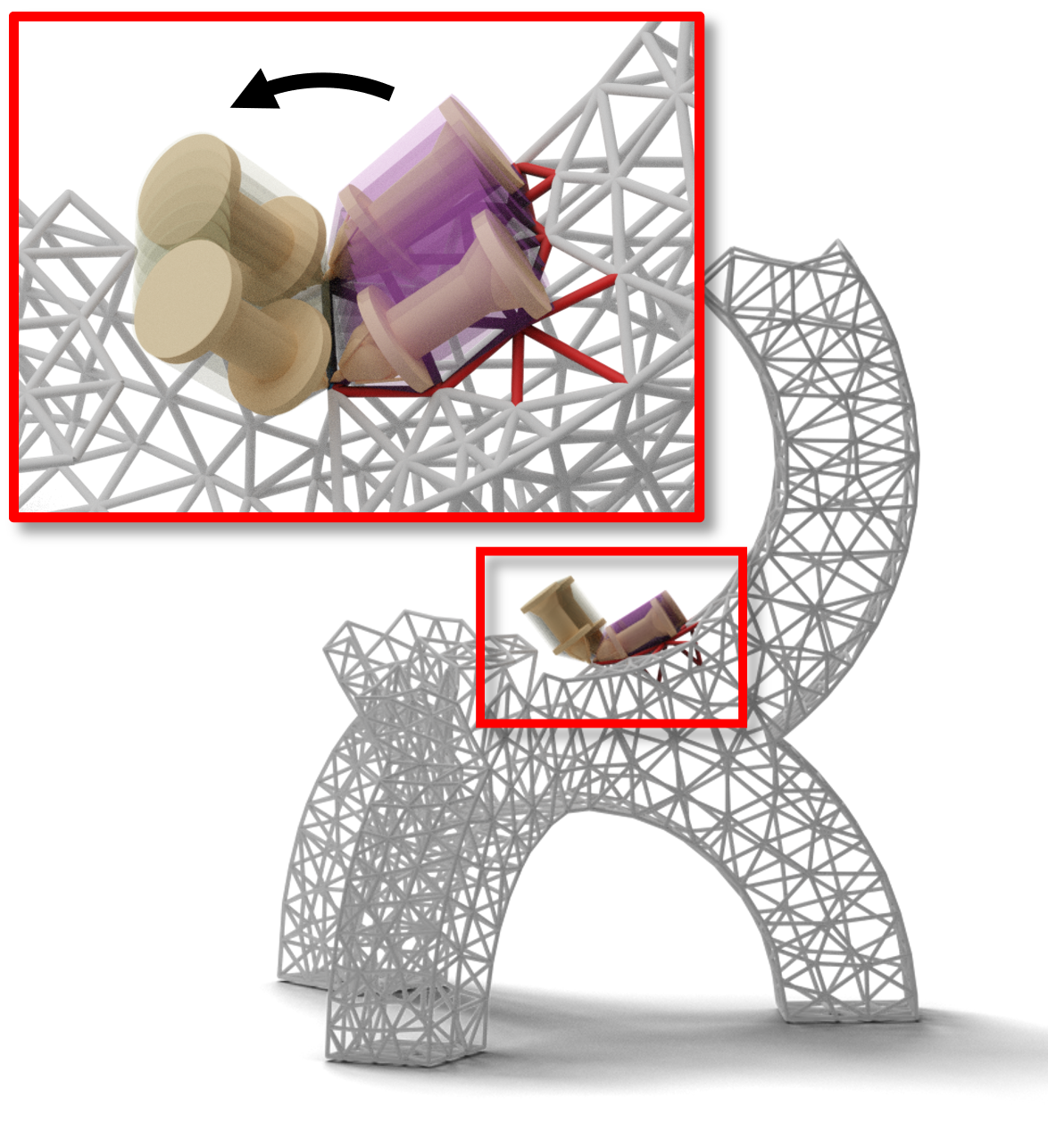}
\end{center}
\end{wrapfigure}

\vspace{5pt}
\subsubsection{Reward for Wire-Frame Printing}~We represent all already printed edges on a wire-frame model as cylinders that can be approximated as a collection of convex polytopes $\mathcal{W}$. 
When attempting to take the printing action to move the printer head from a node $v_a$ to a node $v_b$, we would keep a constant orientation for the printer head (i.e., its orientation $\mathbf{q}_a$ at $v_a$) to simplify its motion, which can be realized on both the 3-axis and the 5-axis printers. 
To compute the swept volume of this motion according to the orientation $\mathbf{q}_a$ -- denoted by $\mathcal{S}(\mathbf{q}_a)$, we simplify the geometry of a printer head by the convex-hull (see Figure on the right for an illustration).
This simplification leads to a convex shape for $\mathcal{S}(\mathbf{q}_a)$. Collision occurs when $\mathcal{W} \cap \mathcal{S}(\mathbf{q}_a) \neq \emptyset$ as shown in light purple region. The collision term of reward function is defined as 
\begin{equation}\label{eq:RewardCollision}
C(v_a,v_b) = 
\left\{
    \begin{array}{ll}
        0, & ~~(\mathcal{W} \cap \mathcal{S}(\mathbf{q}_a) = \emptyset)) \\
        -1000. & ~~(\mathcal{W} \cap \mathcal{S}(\mathbf{q}_a) \neq \emptyset)) \\
    \end{array}
\right.
\end{equation}

When a 5-axis printer (or a robotic arm) is employed, we can adjust the orientation of a printer head by a minimal rotation to find collision-free trajectories. The collision-free orientation $\mathbf{q}_b$ is determined by 
\begin{equation}\label{eq:minRotCollisionFreePose}
\mathbf{q}_b = \arg \min_{\mathbf{q}} \theta(\mathbf{q}_a,\mathbf{q}) \quad s.t., \mathcal{W} \cap \mathcal{S}(\mathbf{q})=\emptyset
\end{equation}
with $\theta(\cdot)$ returning the angle difference (in radian) between two orientations. A sampling based method is employed in our implementation to solve this equation on the Gauss sphere. 
 
By allowing the change of a printer head's orientation, the collision term of our reward function is defined as 
\begin{equation}\label{eq:RewardCollision2}
C(v_a,v_b) = 
\left\{
    \begin{array}{ll}
        -\theta(\mathbf{q}_a,\mathbf{q}_b) & ~~(\mathcal{W} \cap \mathcal{S}(\mathbf{q}_b) = \emptyset)) \\
        -1000 & ~~(\mathcal{W} \cap \mathcal{S}(\mathbf{q}_b) \neq \emptyset)) \\
    \end{array}
\right.
\end{equation}
where $\mathbf{q}_b$ is the solution of Eq.(\ref{eq:minRotCollisionFreePose}). 
The objective of coverage has been considered in the collision term $C(v_a,v_b)$ as collision will be reported when attempting to move the printer head along $v_a v_b$ that has already been printed as a strut in prior actions. 

The other essential part of the reward function is the maximal displacement $U(v_a,v_b)$ on a partially completed wire-frame structure after adding the new strut $v_a v_b$, where the displacements of all nodes are computed by the FEA method proposed in \cite{wang2013cost}. The function of $i$-th step reward in our $Q$-learning based planner for wire-frame printing is then defined as
\begin{equation}\label{eq:RewardWireframe}
    r_i = \Gamma(i) (C(v_{i-1},v_i) + U(v_{i-1},v_i))
\end{equation}
with $v_0 = v_c$ being the center of an LSG.

\begin{figure}
\centering
\includegraphics[width=\linewidth]{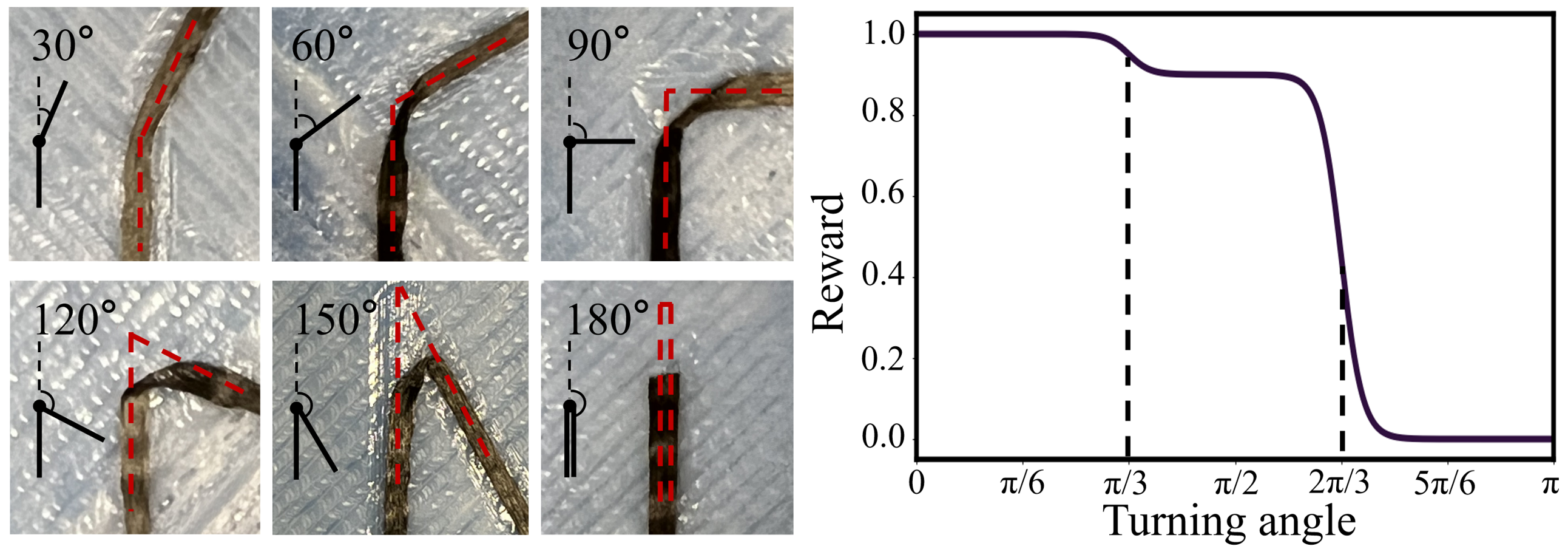}
\caption{(Left) Examples to demonstrate the more and more serious failure cases while increasing the turning angle of CCF toolpath. (Right) The curve of turning-angle reward for CCF printing.}
\label{fig:TurningAngleRewardCurve}
\end{figure}

\vspace{5pt}
\subsubsection{Reward for CCF Printing}~When printing the CCF along the edges of an input graph, one major objective is to control the turning angle $\alpha$ at a node. As can be observed from Fig.~\ref{fig:TurningAngleRewardCurve}, the printed CCF starts to escape away from the planned path when $\alpha$ is larger than $\pi /3$ and becomes unacceptable if $\alpha \in (2 \pi /3, \pi]$. Therefore, we reward all the cases with angle less than $\pi /3$, minimize the angles when $\alpha \in [\pi /3, 2 \pi /3]$ and penalize any angle larger than $ 2 \pi /3$. In short, the following reward of turning-angle is employed.
\begin{equation}
A(v) = \tau\left(\frac{\pi/3 - \alpha(v)}{\pi/60}\right) + 0.9 \left(\tau\left(\frac{\alpha(v) - \pi/3}{\pi/60}\right) - \tau\left(\frac{\alpha(v)-2\pi/3}{\pi/60}\right) \right)
\end{equation}
with $\tau(x) = (1 + \exp({-x}))^{-1}$. Function curve of the above turning-angle reward is shown in the right of Fig.~\ref{fig:TurningAngleRewardCurve}. 

By allowing each edge on an input graph $\mathcal{G}$ to be traveled up to twice, the planning problem of CCF toolpaths has a `bound' in terms of continuity now -- i.e., $\mathcal{G}$ becomes an Euler graph by duplicating every edge. We also did not observe any example that cannot form a continuous toolpath in our experiments. Nevertheless, we still need to minimize the total length of a toolpath to save material usage. Therefore, a negative penalty is given when traveling the edge $(v_a,v_b)$ in the second time. An edge traversal reward is then defined as:
\begin{equation}
D(v_a,v_b) = 
\left\{
    \begin{array}{ll}
        0, & ~~((v_a,v_b) \notin \mathcal{P}) \\
        -L(v_a,v_b)/ L_{\max}, & ~~((v_a,v_b) \in \mathcal{P}) \\
    \end{array}
\right.
\end{equation}
with $L(\cdot,\cdot)$ giving the length of an edge and $L_{\max}$ being the maximal edge length in the input graph $\mathcal{G}$.

The function of $i$-th step reward in our $Q$-learning based planner for CCF printing is then defined as
\begin{equation}\label{eq:RewardCCF}
    r_i = \Gamma(i) (A(v_i) + D(v_{i-1},v_i))
\end{equation}
again with $v_0$ being the center $v_c$ of an LSG.

\vspace{5pt}
\subsubsection{Reward for Metallic Printing}~The area of high temperature region is the major manufacturing objective to be minimized. The temperature on every node in $\mathcal{G}$ is stored as $T(\cdot)$ and updated by the following method during the planning process. Specifically, when moving the laser beam to a node $v_c$, a kernel is applied to $v_c$ by adding the following heat to all nodes $v \in \mathcal{G}$
\begin{equation}
H(v) = 
\left\{
    \begin{array}{ll}
        H_{\max}(1-(d(v,v_c)/R)^{0.3}), & ~~(d(v,v_c)< R) \\
        0, & ~~(d(v,v_c) \geq R) \\
    \end{array}
\right.
\end{equation}
with $d(\cdot,\cdot)$ denoting the distance between two nodes. $H_{\max}$ is the maximal heat generated by the laser beam. $R = 3 \bar{L}$ with $\bar{L}$ being the average edge length on a graph. 

For each action of move, we first impose the above heat source as $T(v)=T(v)+H(v)$ and then update the temperature on every node by solving a diffusion equation. For an efficient implementation, we apply a local Laplacian operator with the support range as $2n$ rings. With the updated temperature distribution,  the reward function of $i$-th step reward in our $Q$-learning planner for metallic printing can be defined as 
\begin{equation}
r_i = 
\left\{
    \begin{array}{ll}
        -\Gamma(i) T(v_i) , & ~~(v_i \notin \mathcal{P}) \\
        -1000, & ~~(v_i \in \mathcal{P}) \\
    \end{array}
\right.
\end{equation}
where the second part is to penalize the multiple visits of a node. Here the reward function drives the planner to generate toolpaths travelling into the cooler regions.

\section{Implementation details}\label{secDetails}
\begin{figure}
\centering
\includegraphics[width=\linewidth]{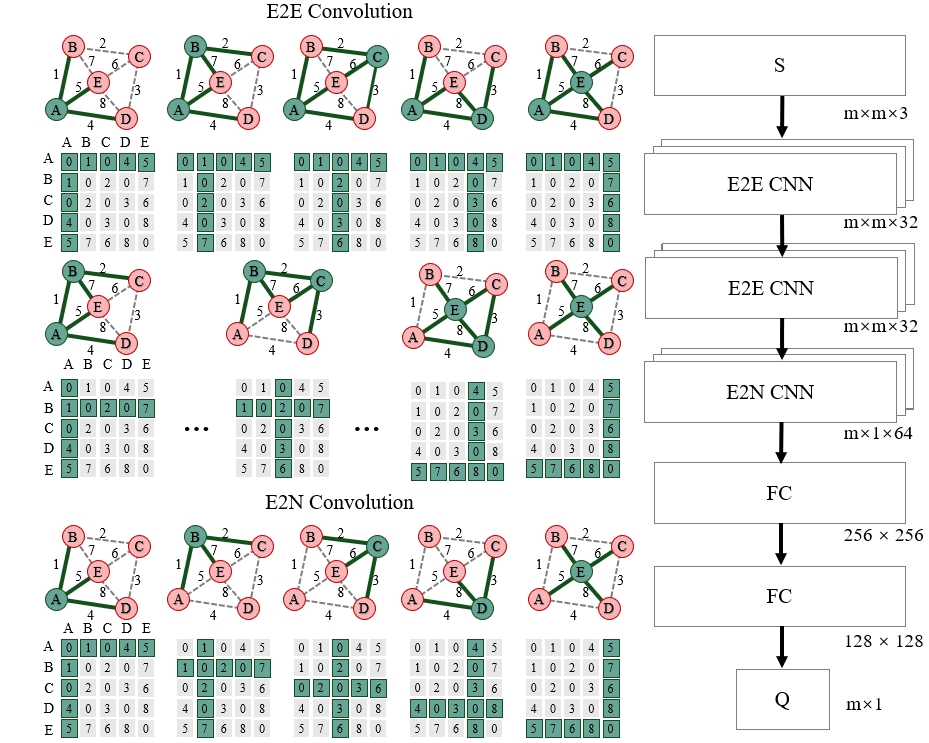}
\put(-243,3){\small \color{black}(a)}
\put(-73,3){\small \color{black}(b)}
\caption{The illustration of (a) the operators for the \textit{Edge-to-Edge} (E2E) and \textit{Edge-to-Node} (E2N) convolution layers, where the relevant nodes / edges and the elements in the adjacent matrix are highlighted by the green color. (b) The neural network architecture employed in our DQN-based learning.
}\label{fig:CNN}
\end{figure}
\subsection{Convolution operator}
$Q$-learning algorithm usually employed the \textit{convolutional neural network} (CNN) based on standard 2D convolutions, which primarily focus on spatial locality in image-like data. To capture the topological information on the input as the adjacency matrix of a graph, we seek the help from the row-column convolution operators (e.g.,~\cite{KAWAHARA20171038}) that can effectively transmit information between two nodes even if they are located away from each other in the adjacent matrix. Specifically, \textit{Edge-to-Edge} (E2E) and \textit{Edge-to-Node} (E2N) convolutions are employed. For a matrix $\mathbf{A}_{m \times m}$ that represents a LSG, the E2E convolution is applied to every elements $a_{i,j} \in \mathbf{A}$ by the weights on the same row and the same column of a coefficient layer $\mathbf{W}=[w_{i,j}]_{m \times m}$ as
\begin{equation}\label{eq:E2EConvOperator}
    a^{*}_{i,j} = \sum_{k=1}^m w_{i,k} a_{i,k} + \sum_{k=1}^m w_{k, j} a_{k, j}.
\end{equation}
Differently, the E2N convolution is only applied to the diagonal elements $a_{i,i} \in \mathbf{A}$ by the weights on the same row and the same column of a coefficient layer $\mathbf{W}$ as
\begin{equation}\label{eq:E2NConvOperator}
    a^{*}_{i,i} = \sum_{k=1}^m w_{i,k} a_{i,k} + \sum_{k=1}^m w_{k, i} a_{k, i}.
\end{equation}
As illustrated in Fig.\ref{fig:CNN}(a), the relevant nodes, edges and weights of the convolution applied to different $a_{i,j}$ are highlighted by green color. More details of these convolution operators can be found in \cite{KAWAHARA20171038}.

\subsection{Network architecture}\label{subsecDetailNetwork}
The network of DQN employed in our $Q$-learning based planner contains five layers in total. The major network parameters are stored in two layers of E2E convolutions, one layer of E2N convolutions, and two fully connected layers -- see Fig.~\ref{fig:CNN}(b). The input of the network is the moving state $\mathbf{S}^*_{m \times m \times 3}$ of a LSG, and the output is a vector of $Q$-values for all nodes. These layers, by integrating edge and node perspectives, enable our planner to effectively handle complex network structures, broadening their applicability to non-grid data (i.e., the diverse graphs for 3D printing). More studies for the network architecture can be found in the supplementary document.

\begin{figure}
\centering
\includegraphics[width=\linewidth]{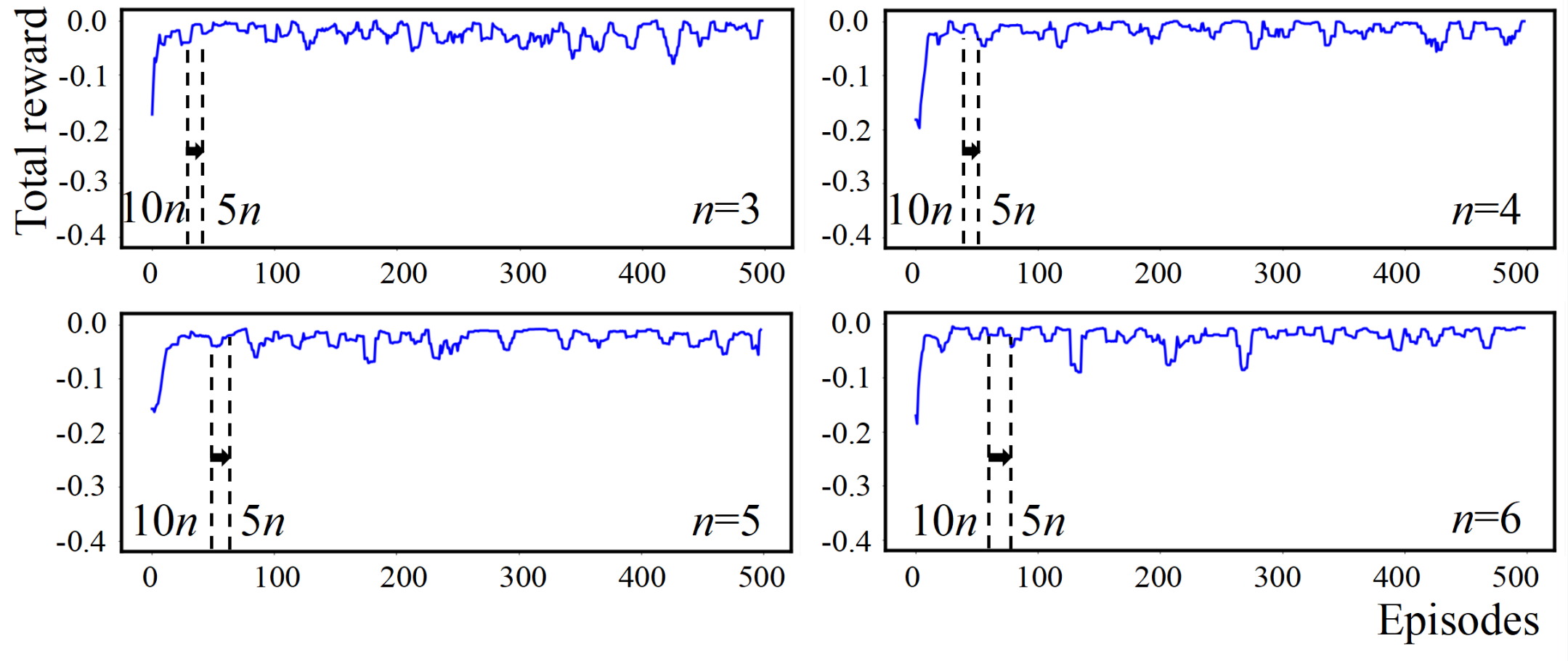}
\caption{The study of required numbers of episodes for the reinforcement learning to converge by using different range size $n$ of an LSG tested on the Bunny model in Fig.~\ref{fig:Teaser}, where the total reward is evaluated as $\sum_{i=1}^n r_i$. 
}\label{fig:convergenceCurve}
\end{figure}

\subsection{Termination criteria of learning}\label{subsecTerminalCond}
For the termination criterion used in \rev{the learning routine of} the planner, we first study the required number of episodes for the learning process to converge by using different range size $n$ of an LSG. As shown in Fig.~\ref{fig:convergenceCurve} for the tests taken on the Bunny model, the learning process can be empirically considered as converged when the best total reward $\sum_{i=1}^n r_i$ among all tested paths has not changed after $5n$ episodes, which often occurs after $10n$ episodes. Therefore, we define the following hybrid termination criteria for the `optimization' process of our planner:
\begin{enumerate}
\item The learning has been conducted for more than $10n$ episodes and the best total reward has not changed in the past $5n$ episodes; or 
\item The learning has been conducted for more than 500 episodes.
\end{enumerate}
The second criterion is employed to control the maximally allowed time for learning. 

We need also to consider the termination criterion for the toolpath generation algorithm, which is actually defined according to different 3D printing applications -- i.e., mainly by the coverage requirements. Planners for the wire-frame printing and the CCF printing stop when all edges have been added into the toolpath $\mathcal{P}$. Differently, the planner for metallic printing stops after all nodes have been included in $\mathcal{G}$. 

\begin{figure}
\centering
\includegraphics[width=\linewidth]{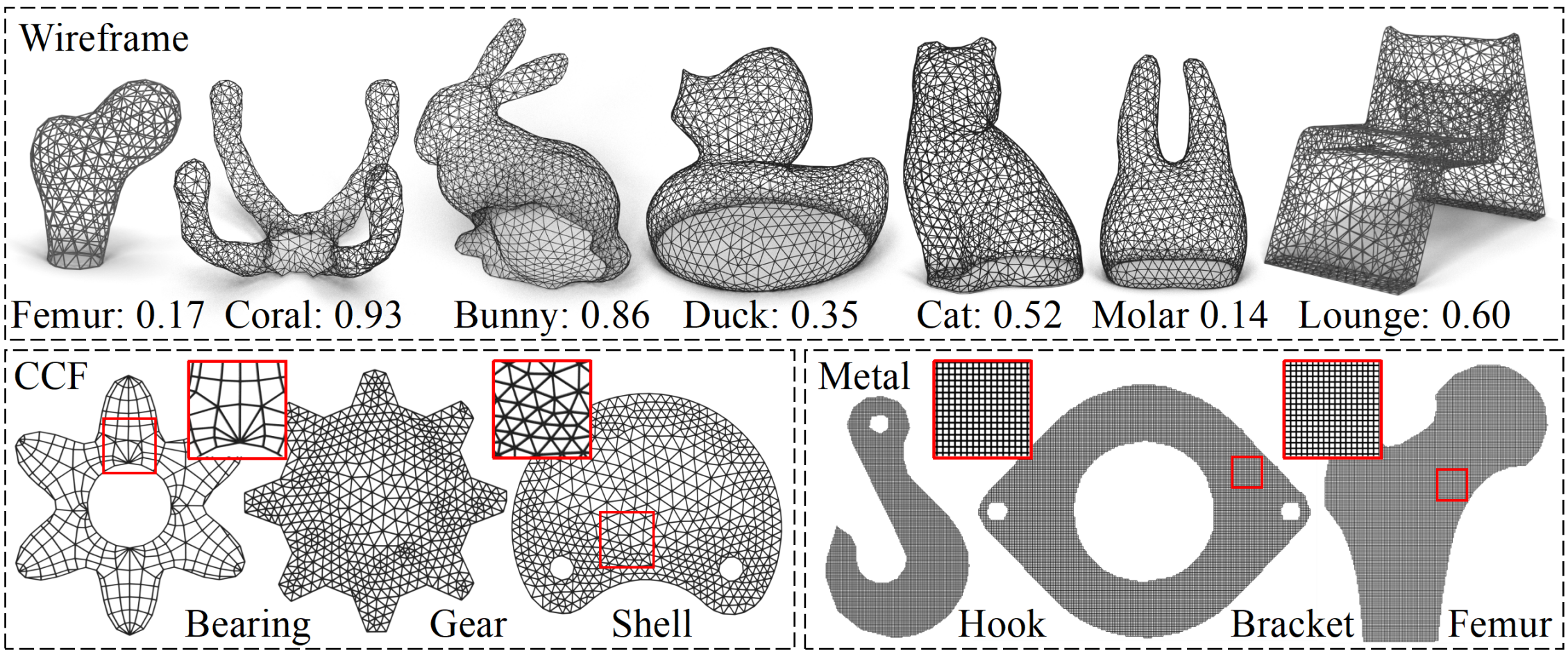}
\caption{The performance of our learning-based planner has been tested on a variety of models for (a) wire-frame printing, (b) CCF printing and (c) metallic printing, where the maximal displacements $U_{\max}$ from the planning results are given for the wire-frame printing examples.}
\label{fig:allModels}
\end{figure}

\subsection{Acceleration for collision avoidance}
$C(v_{i-1},v_{i})$ is employed to compute the reward in Eq.(\ref{eq:RewardCollision2}) for collision avoidance. This can be further accelerated by directly assigning $C(v_{i-1},v_{i})=-1000$ when collision occurs on any previous edge $(v_{j-1},v_{j})$ with $j<i$ on the planned path. According to our experiments, this acceleration has no influence on the quality of the resultant toolpaths but can achieve 3-5 times speed up.

\section{Results and Verification}\label{secResult}
\subsection{Computational Results}
We have implemented our learning-based planner in Python. The computational experiments are all conducted on a PC with Intel Core i7-13700F CPU and RTX 4080 GPU. Our method has been tested on a variety of models (as shown in Fig.~\ref{fig:allModels}), where the computational statistics are given in Table.~\ref{tab:compStatisticsRes}. 
Optimized toolpaths can be successfully planned on models with up to $39k$ edges, which demonstrates the scalability of our planner. In these examples, the demanded manufacturing objectives in the 3D printing applications of wire-frame models, CFRTPs, and LPBF-based metallic objects are considered. 

\begin{table}
\caption{The computational statistics for different testing models.}
\centering\label{tab:compStatisticsRes}
\footnotesize
\vspace{-5pt}
\begin{tabular}{r|c||r|r||c||r|r}
\hline
 & & \multicolumn{2}{c||}{Input Graph} & Range & \multicolumn{2}{c}{Comp. Time (min.)} \\
\cline{3-4}\cline{6-7}
Model & Type & Node \# & Edge \# & of LSG & Primary & Reuse$^\dag$ \\
\hline
\hline
Femur &  &  263 & 772 & $n=6$ & 38.11 & 22.38\\
Coral &  &  785 & 2,322 & $n=6$ & 116.84 & 75.7\\
Cat &  &  823 & 2,412 & $n=6$ & 125.22 & 86.81\\ 
Molar & Wire-frame &  852 & 2,486 & $n=6$ & 138.78 & 90.91\\ 
Bunny &  & 1,155 & 3,382 & $n=6$ & 188.55 & 122.96\\
Duck &  & 1,193 & 3,476 & $n=6$ & 203.27 & 130.62\\
Lounge &  & 1,390 & 4,164 & $n=6$ & 235.14 & 160.82\\
\hline
Bearing &  & 294 & 575 & $n=6$ & 39.33 & 28.09\\  
Gear & CCF & 551 & 1,550 & $n=6$ &  100.12 & 58.15\\ 
Shell &  & 567 & 1,599 & $n=6$ &  104.35 & 59.35\\ 
\hline
Hook &  & 13,038 & 25,604 & $n=6$ & 766.65 & 505.38\\
Bracket & Metal & 15,194 & 29,814 & $n=6$ & 958.86 & 651.93\\
Femur &  & 19,858 & 39,301 & $n=6$ & 1290.06 & 854.22\\
\hline
\end{tabular}
\begin{flushleft}
$^\dag$~Refers to the computing time of the accelerated scheme with prior reuse.
\end{flushleft}
\end{table}  

\begin{figure}
\centering
\includegraphics[width=\linewidth]{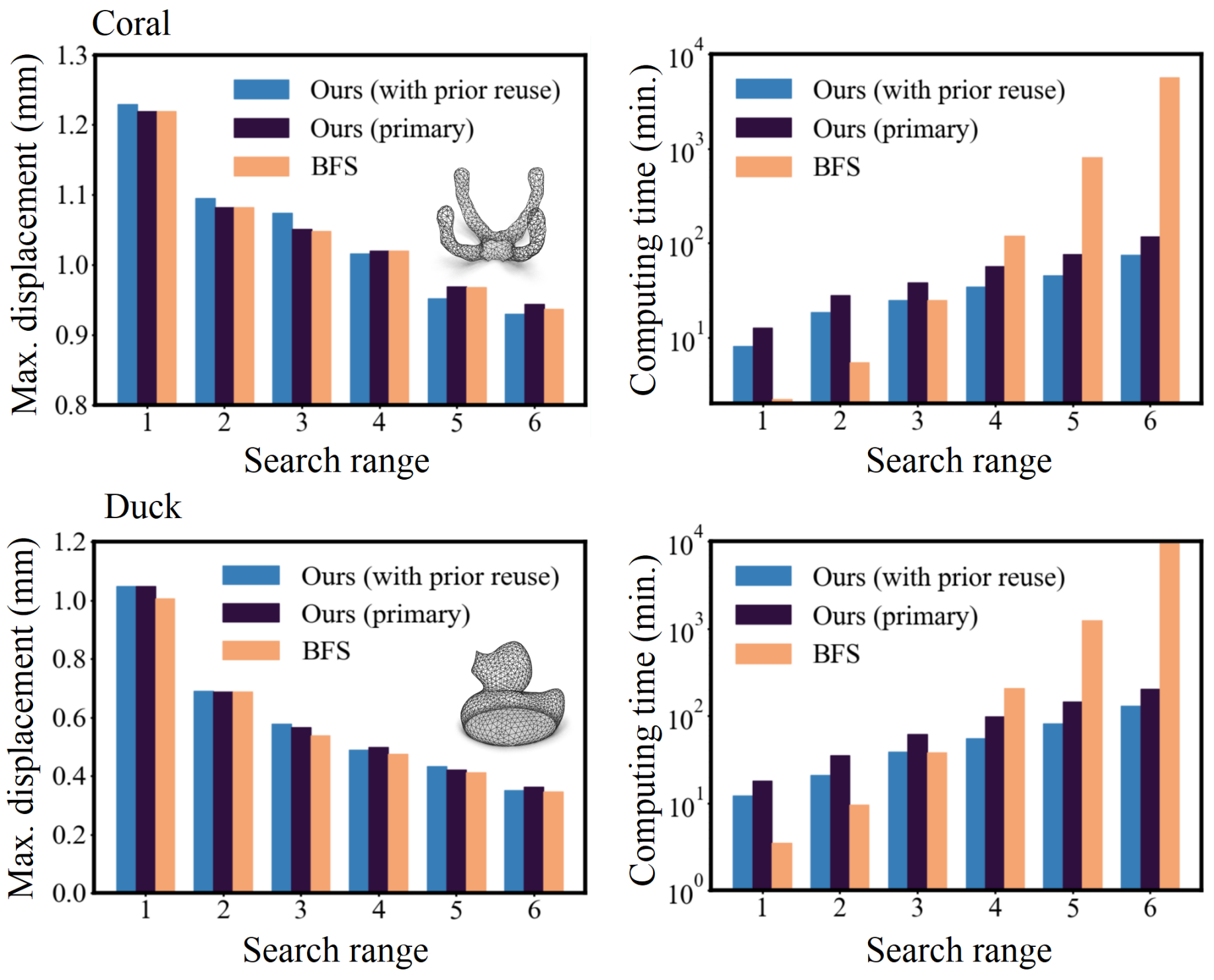}
\put(-242,102){\small \color{black}(a)}
\put(-242,4){\small \color{black}(b)}
\vspace{-5pt}
\caption{The study of performance and computing time for generating wire-frame printing toolpaths on (a) the Coral model and (b) the Cat model.
}\label{fig:comparisonBFSCoralCat}
\end{figure}

We first consider the requirement of minimized deformation to be achieved in wire-frame printing. \rev{Beam-based FEA is employed in our implementation, where the FEA takes about 1.2ms in average for a LSG with $n=6$.} The first example is the Bunny model the result of which has been given in Fig.~\ref{fig:Teaser}. The maximal displacement of the partially printed structures by our toolpath is $0.86\mathrm{mm}$, where the diameter of struts is $d=1.00\mathrm{mm}$ and the material properties are $E=3.84\mathrm{GPa}$ for Young's modulus and $\nu = 0.35$ for Poisson's ratio. Similarly, the deformation represented as the maximal displacement is $0.94\mathrm{mm}$ on the Coral model (Fig.\ref{fig:comparisonBFSCoralCat}(a)) and $0.35\mathrm{mm}$ on the Duck model (Fig.\ref{fig:comparisonBFSCoralCat}(b)) with $d=1.00\mathrm{mm}$, $E=2.64\mathrm{GPa}$ and $\nu = 0.35$. The maximal displacements $U_{\max}$ on other models are given in Fig.\ref{fig:allModels}. 

We then study the influence of an LSG's range $n$ by comparing our planner with the BFS algorithm in terms of time and performance. As shown in Fig.~\ref{fig:Teaser} and Fig.~\ref{fig:comparisonBFSCoralCat}, when $n \leq 4$ the computation time of BFS algorithm is similar to or slightly shorter than our planner. However, the performance of planning by using such a small range of monitoring is usually less optimal -- i.e., with displacement larger than half of the nozzle's diameter as $1.0\mathrm{mm}$. 
When increasing to $n=5$, the time of BFS algorithm raises to $15.50 \times$ of our planner on the Bunny model (Fig.~\ref{fig:Teaser}), $17.86 \times$ on the Coral model (Fig.~\ref{fig:comparisonBFSCoralCat}(a)) and $15.28 \times$ on the Duck model (Fig.~\ref{fig:comparisonBFSCoralCat}(b)). The computing time of the BFS algorithm increases exponentially when enlarging the range of LSGs -- which can be more than $214$ hours for $n=6$ on the Bunny model. For the same case, the computing time of our method increases linearly and needs less than $2.05$ hours. Similar performance can be found in CCF printing (see supplementary document for the details). In summary, our learning based planner is scalable to plan toolpaths on graphs with large number of nodes / edges. 


\begin{figure}
\centering
\includegraphics[width=\linewidth]{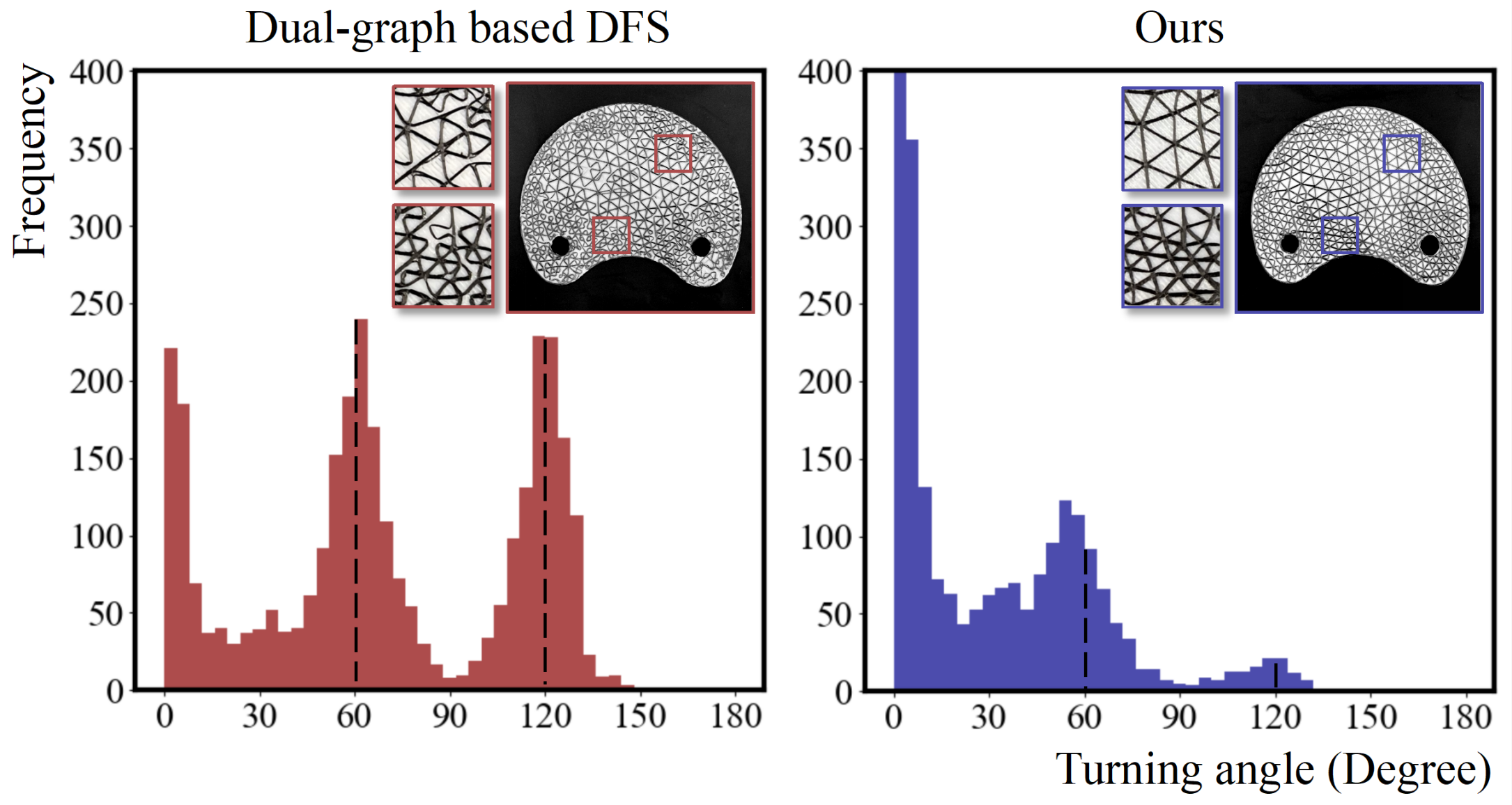}
\vspace{-15pt}
\caption{The illustration of the turning angle distribution on the Shell model for the dual-graph based DFS method \cite{huang2023turning} and our path. The comparison of the two figures shows that not only turning angles greater than $120^{\circ}$ are clearly optimized by our path with less frequency of distribution, but also turning angles between $60^{\circ}$ and $120^{\circ}$ are optimized.
}\label{fig:angle distribution}
\end{figure}

\begin{figure}
\centering
\includegraphics[width=\linewidth]{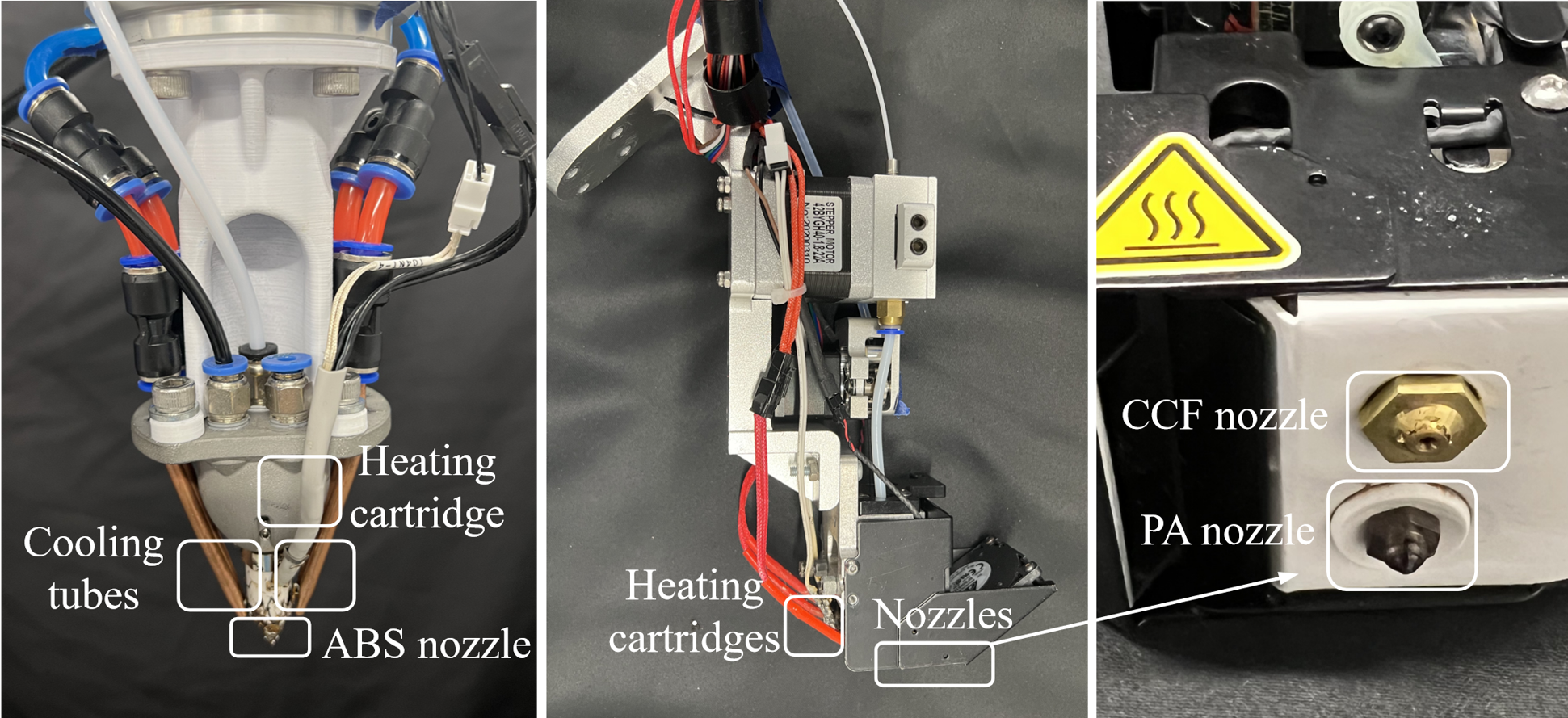}
\put(-241,4){\small \color{white}(a)}
\put(-158,4){\small \color{white}(b)}
\caption{Hardware used in our physical experiments for (a) wire-frame printing -- the printing head and (b) CCF printing -- the printer head.}\label{fig:hardwareCCF_and_Wireframe}
\end{figure}

\begin{figure*}
\centering
\includegraphics[width=\linewidth]{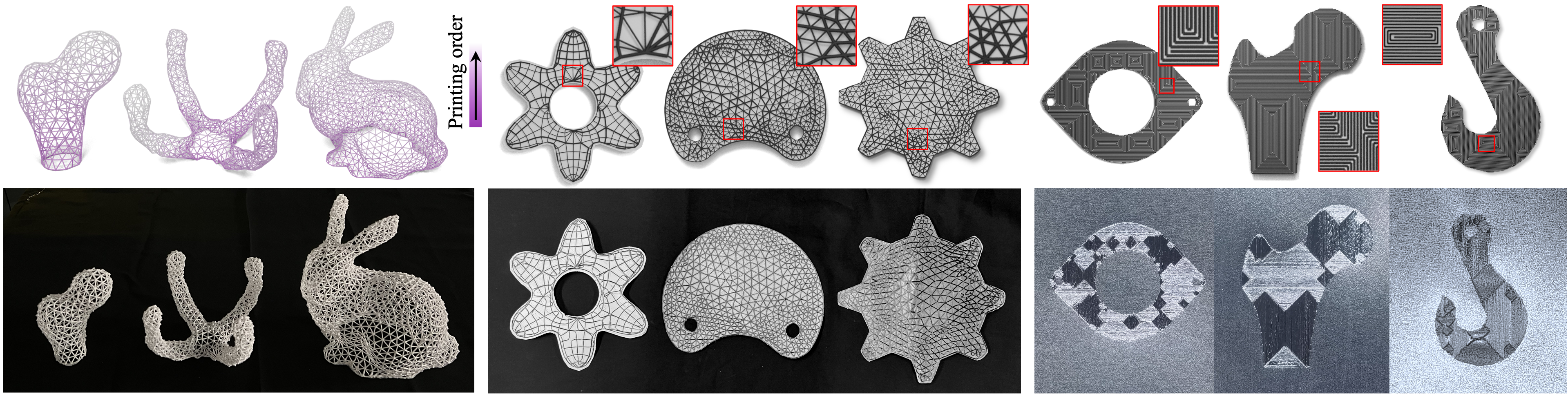}
\put(-509,120){\small (a)}
\put(-353,120){\small (b)}
\put(-174,120){\small (c)}
\caption{Our method has been tested on a variety of models by incorporating the manufacturing objectives in different applications: (a) wire-frame printing with colors visualizing the order of printing, (b) CCF printing for CFRTPs, and (c) LPBF-based metallic printing. (Top Row) Computational results for toolpath planning. (Bottom Row) Fabrication results for the corresponding models.}\label{fig:fabResults}
\end{figure*}

For the quality of toolpaths generated for CCF printing, we have tested and verified on the four models shown in Fig.~\ref{fig:allModels}. The results generated by our learning-based planner are quite encouraging -- see Table.~\ref{tab:CCFToolpath}. Comparing to the most recently published method presented in \cite{huang2023turning} that is a DFS method with backtracking, our method can significantly reduce the number of sharp turns (i.e., those with turning angles larger than $120^{\circ}$) and the total length of the toolpath, which are two major manufacturing objectives to be achieved for CCF printing. Meanwhile, turning angles between $60^{\circ}$ and $120^{\circ}$ have also been optimized as shown in Fig.~\ref{fig:angle distribution}.

\begin{table}
\caption{Statistics for evaluating the quality of CCF toolpath.}
\centering\label{tab:CCFToolpath}
\footnotesize
\vspace{-5pt}
\begin{tabular}{r||c|r||c|r}
\hline
 &  \multicolumn{2}{c||}{\# of sharp turns (i.e., $\geq 120^{\circ}$)}  & \multicolumn{2}{c}{Length of toolpath (meter))}  \\
\cline{2-3}\cline{4-5}
Models & DFS method$^\dag$ & Our method & DFS method$^\dag$ & Our method \\
\hline \hline
Bearing         & $31$  & $8$~($\downarrow 74.19\%$) & $12.49$ &  $9.04$~($\downarrow 27.62\%$)     \\ 
Gear         & $626$  &  $44$~($\downarrow 92.97\%$)  & $29.46$ & $21.10$~($\downarrow 28.38\%$)    \\ 
Shell         & $730$  &  $49$~($\downarrow 93.29\%$)  & $31.06$ & $22.51$~($\downarrow 27.53\%$)    \\ 
\hline
\end{tabular}
\begin{flushleft}
$^\dag$~The previous paths compared here are generated by the dual-graph based DFS method (ref.~\cite{huang2023turning}). 
\end{flushleft}
\end{table}  

\subsection{Hardware and Fabrication Parameters}
We have tested the toolpaths generated by our learning-based planner in physical experiments, which are conducted on three different hardware systems as introduced below.

\subsubsection{Wire-frame printing}
The system consists of a printer head with a 1.0mm nozzle (see Fig.\ref{fig:hardwareCCF_and_Wireframe}(a)), a cooling mechanism of four copper tubes with strong air-flow provided by a pump and a UR5e robotic arm with repeatability at $\pm 0.03\mathrm{mm}$ to provide 6-DOFs motion (see the right of Fig.~\ref{fig:Teaser} and Fig.~\ref{fig:Three3DPntProblems}(a.2)). Acrylonitrile butadiene styrene (ABS) filaments with diameter $1.75\mathrm{mm}$ are employed in our experiments of wire-frame printing.

\subsubsection{CCF printing}
The system has a dual-material printer head equipped with two parallel printing nozzles. A  nozzle with diameter $1.0\mathrm{mm}$ is used for printing Polyamide (PA) filament with diameter $1.75\mathrm{mm}$ as the matrix material of continuous fiber-reinforced thermoplastic composites (CFRTPCs), and a $2.0\mathrm{mm}$ flat nozzle is employed for printing CCF filament with diameter $0.4\mathrm{mm}$. Each nozzle has a separated extrusion control system, including an extruder, a heating cartridge, and a temperature sensor (see Fig.~\ref{fig:hardwareCCF_and_Wireframe}(b)). The motion is again provided by the UR5e robotic arm (see Fig.~\ref{fig:Three3DPntProblems}(b.2)). Example models are printed with the height as $3.0\mathrm{mm}$ (i.e., $3$ layers of material and $3$ layer of CCF toolpaths).

\subsubsection{Metallic printing}
The experiments of metallic printing are conducted on an LPBF machine (see Fig.~\ref{fig:Three3DPntProblems}(c.2)) featured with a laser beam at the size of 50 micrometers to print 316L stainless steel. Example models are printed at the height of $0.2\mathrm{mm}$ (i.e., one layer).

\begin{figure}   
\centering
\includegraphics[width=\linewidth]{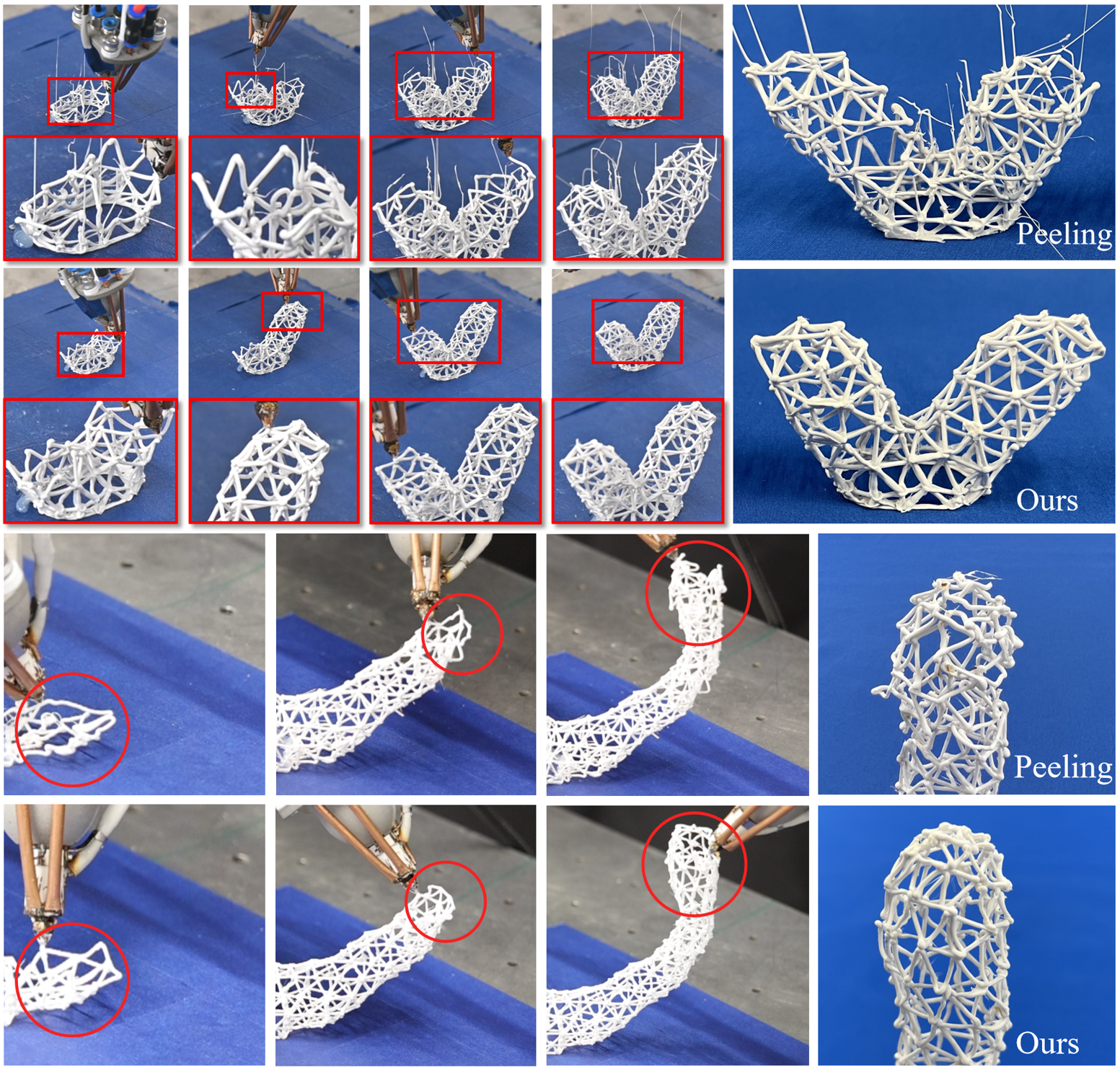}
\caption{Comparison of the printing results of two models by using the toolpath generated by peeling heuristic \cite{wu2016printing} vs.\ the toolpath generated by our learning-based planner. When using the peeling-like toolpath, we can observe the failure cases of 1) large bending deformation on struts and 2) failed strut printing caused by misaligned positions of endpoints -- i.e., with large displacements caused by gravity. Also, stringing phenomenon occurs due to frequent lifting of the tool (see the top row).
}\label{fig:failureWireframePeeling}
\end{figure}

\subsection{Verification by Physical Fabrication}
The performance of toolpaths generated by our method has been verified by physical fabrication, where the results of example models can be found in Fig.~\ref{fig:fabResults}. The progressive printing procedure has also been demonstrated in the supplementary video.

\begin{figure}
\centering
\includegraphics[width=\linewidth]{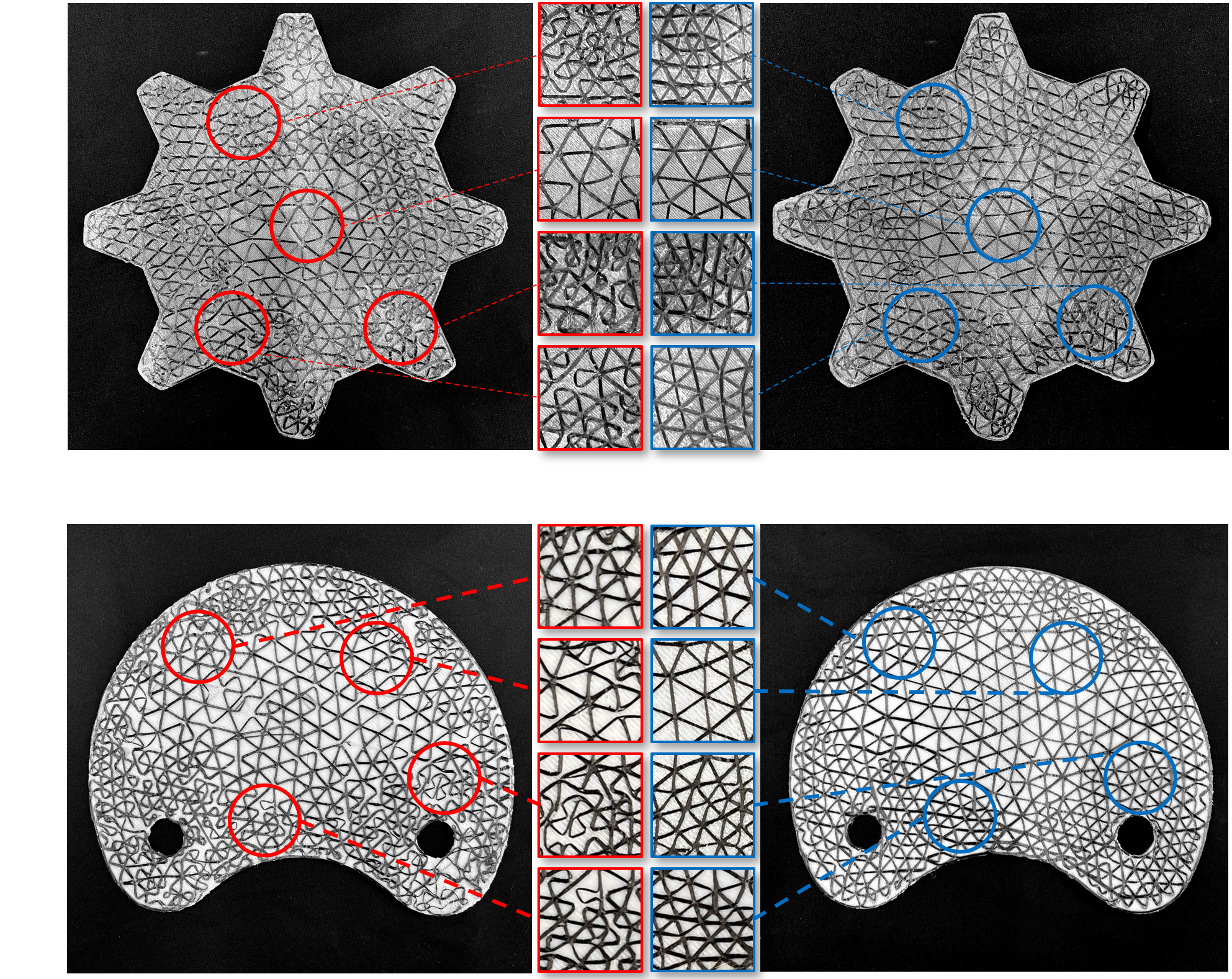}
\put(-242,97){\small \color{black}(a)}
\put(-230,97){\small \color{black}\# Sharp turns: 626}
\put(-90,97){\small \color{black}\# Sharp turns: 44}
\put(-242,-6){\small \color{black}(b)}
\put(-230,-6){\small \color{black}\# Sharp turns: 730}
\put(-90,-6){\small \color{black}\# Sharp turns: 49}
\caption{The number of sharp turns (i.e., with turning angle $\geq 120^{\circ}$) can be significantly reduced by our learning-based planner: (a) Gear model -- the number of sharp turns is reduced from 626 to 44 (reduced by $92.97\%$) and (b) Shell model -- the number of sharp turns is reduced from 730 to 49 (reduced by $93.29\%$). For both models, we highlight four regions to demonstrate the improvement of printing quality when less sharp turns are formed -- i.e., the printed fibers give better alignment with the input graph.}
\label{fig:fabResCCF}
\end{figure}

\begin{figure}
\centering
\includegraphics[width=\linewidth]{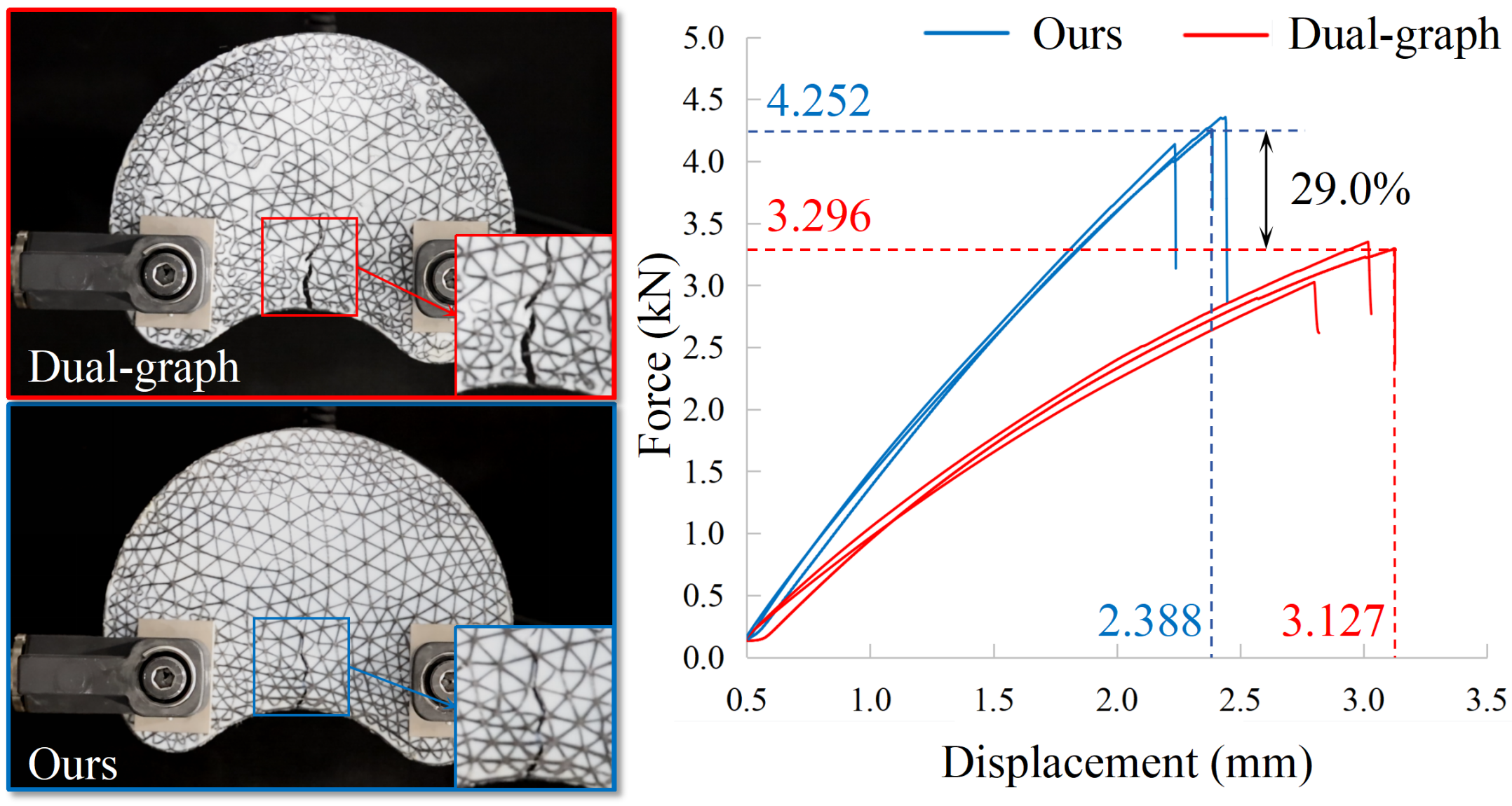}
\caption{
Comparisons of mechanical reinforcement by CCF using different toolpaths for the Shell model, including the toolpath generated by the dual-graph based DFS method \cite{huang2023turning} and our toolpath. As the number of sharp turns have been significantly reduced on our toolpath, stronger mechanical strength is observed in the tensile tests. Specifically, specimen fabricated by our toolpath shows 29.00\% increased breaking force while using 27.53\% less CCF. 
}\label{fig:fabResCCF_TensileTest}
\end{figure}

\begin{figure}
\centering
\includegraphics[width=\linewidth]{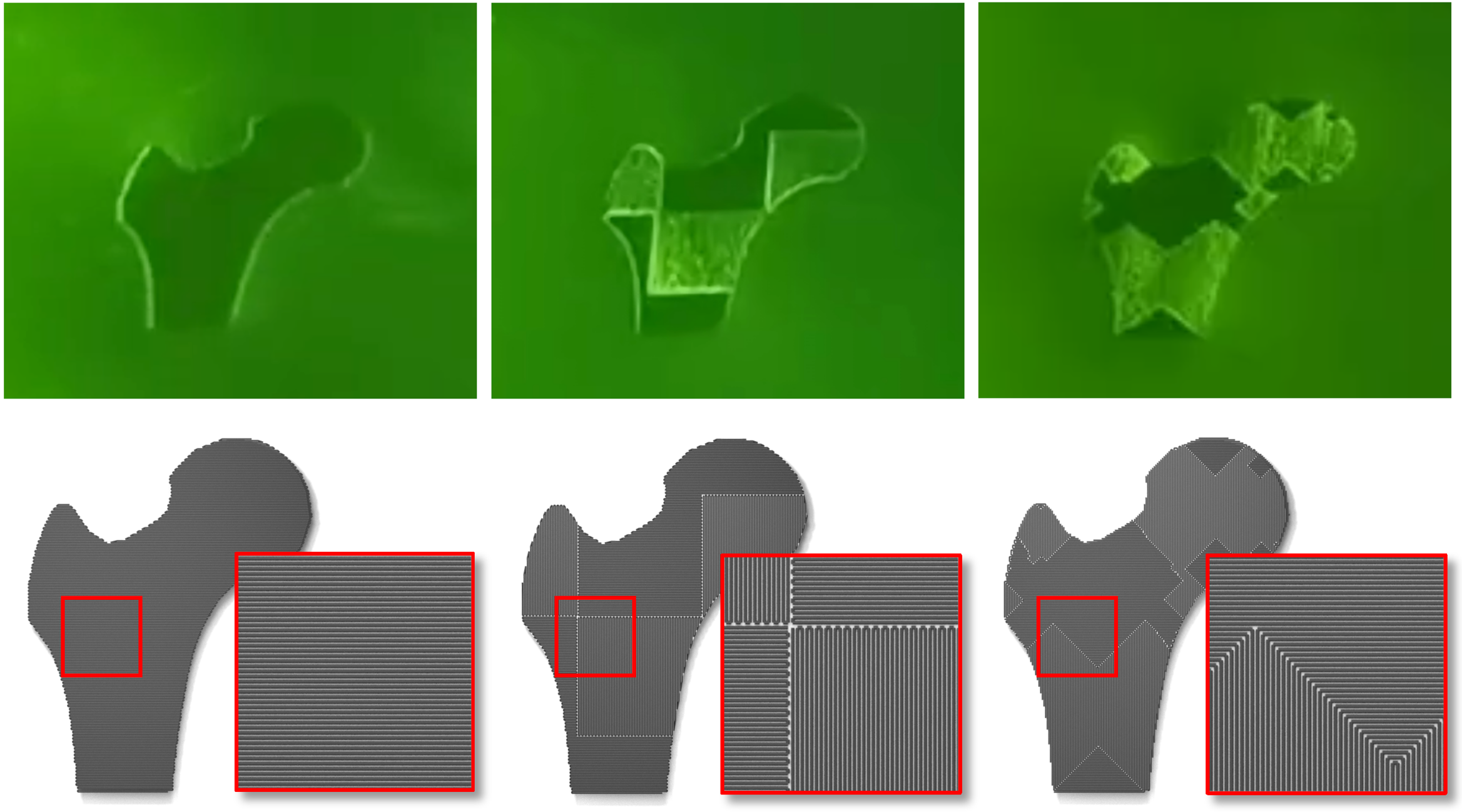}
\put(-213,72){\small \color{white}Zigzag}
\put(-141,72){\small \color{white}Chessboard}
\put(-50,72){\small \color{white}Ours}
\caption{LPBF processes of metallic printing by using the zigzag toolpath, the chessboard toolpath \cite{qiu2013microstructure}, and our toolpath.}\label{fig:LPBFProcesses}
\end{figure}

We first study the results of wire-frame printing. As shown in Fig.~\ref{fig:Teaser} and Fig.~\ref{fig:fabResults}(a), different wire-frame models with up to 3k struts can be successfully printed. This is benefited from the optimized toolpaths that minimize the deformation caused by gravity on the partially printed structures. The failure fabrication by using unoptimized toolpaths can be found in Fig.~\ref{fig:failureWireframePeeling}, where heuristics presented in \cite{wu2016printing} were employed.

The results of printed CFRTPCs demonstrate the effectiveness of our planner in minimizing the number of sharp turns. As shown in Fig.~\ref{fig:fabResCCF}, toolpaths generated by our method can successfully avoid most of the sharp-turn cases ($\geq 120^{\circ}$) while reducing the CCF consumption. Problems caused by sharp turns can be clearly observed from the zoom views. CCF printing with sharp turns will significantly reduce the effectiveness of mechanical reinforcement that is only introduced along the axial direction of fibers. We further verify this by conducting the tensile tests on the specimens of the Shell model fabricated by using the toolpaths of the dual-graph based DFS method \cite{huang2023turning} and ours. As can be found from the force-displacement curves in Fig.~\ref{fig:fabResCCF_TensileTest}, the breaking force is given on the specimens fabricated by our toolpath is 29.00\% larger while the consumed CCF filaments is 27.53\% less (see Table \ref{tab:CCFToolpath}). Note that we printed the matrix layers of all these CFRTPCs by using the conventional zigzag toolpaths. 

\begin{figure}
\centering
\includegraphics[width=\linewidth]{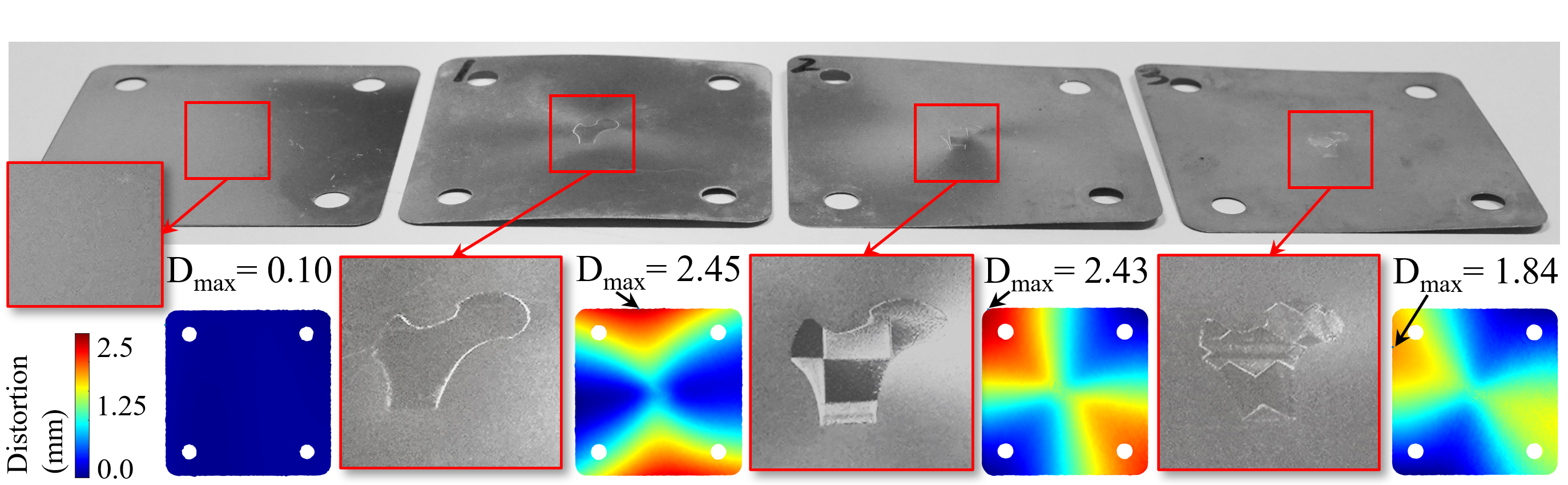}
\put(-223,72){\small \color{black}Flat Plate}
\put(-164,72){\small \color{black}Zigzag}
\put(-114,72){\small \color{black}Chessboard}
\put(-45,72){\small \color{black}Ours}
\caption{Thermal warpage study using different toolpaths to print the Femur model on the initial plate, including the zigzag toolpath, the chessboard toolpath \cite{qiu2013microstructure} and our toolpath. Distortions are measured by 3D scanner and displayed as colormaps. Distortion on the initial plate is $0.10\mathrm{mm}$. Maximum distortion on the specimen printed by our toolpath is $1.84\mathrm{mm}$, which is reduced by 24.90\% and 24.28\% compared to the zigzag toolpath ($2.45\mathrm{mm}$) and the chessboard toolpath ($2.43\mathrm{mm}$).}\label{fig:resMetalDistortion}
\end{figure}

Lastly, we verify the performance of our toolpath in the application of metallic printing to reduce warpage on the resultant models, where the LPBF processes using different toolpaths are shown in Fig.~\ref{fig:LPBFProcesses}. A standard method for evaluating warpage in metallic 3D printing is to print a single-layer model on thin-shell plates with the standard dimension $70\mathrm{mm} \times 70\mathrm{mm}$ (ref.~\cite{ramani2022smartscan,wolfer2019fast, qin2023deep, boissier2022time}). The steel plates we used were made by conventional flattening and thinning with their initial distortion controlled within $0.1 \mathrm{mm}$. Our specimens were printed in a range of $9\mathrm{mm} \times 11\mathrm{mm}$ at the center of the plate. During the printing process, four corners of a plate were fixed by bolts. After printing the single-layer model, the plate was released from the bolts and cooled down for $30$ minutes. We use a KSCAN-Magic 3D scanner to measure the geometry of the plate and display its distortion from the flat plate by color maps (see Fig.~\ref{fig:resMetalDistortion}). It can be observed that the maximum distortion of our result ($1.84\mathrm{mm}$) has decreased by 24.90\% comparing to the zigzag toolpath ($2.45\mathrm{mm}$) and 24.28\%  comparing to the chessboard toolpath ($2.43\mathrm{mm}$). This is because that more dispersed laser energy density was generated by our toolpath.

\section{Discussion and Limitations}\label{secDiscussion}

\subsection{Graph neural network}
To capture the topological information of LSG encoded in the adjacent matrix, we employed the row-column convolution operators in our learning based planner. It is an interesting study to explore if our learning-based planner can be realized by using the recently popular \textit{Graph Convolutional Network} (GCN) that is usually considered as a possible strategy to capture topological information. Specifically, we implemented GCN~\cite{kipf2017semisupervised} by using the common benchmark  library PyG \cite{Fey/Lenssen/2019}. Experiments have been conducted on the Bunny model (Fig.\ref{fig:Teaser}) by LSGs from $n=1$ to $6$. Our method demonstrates deformation similar to GCN-based $Q$-learning implementation (i.e., Ours with the maximal displacement as $U_{\max} = 0.86$ and the result of GCN having $U_{\max} = 0.88$ when $n=6$). 
Moreover, our methods are 19\% - 30\% more efficient due to the more effective memory visit in the matrix-based implementation. Detailed data on more comparisons can be found in the supplementary document.

\subsection{Starting node}
\begin{figure}
\centering
\includegraphics[width=\linewidth]{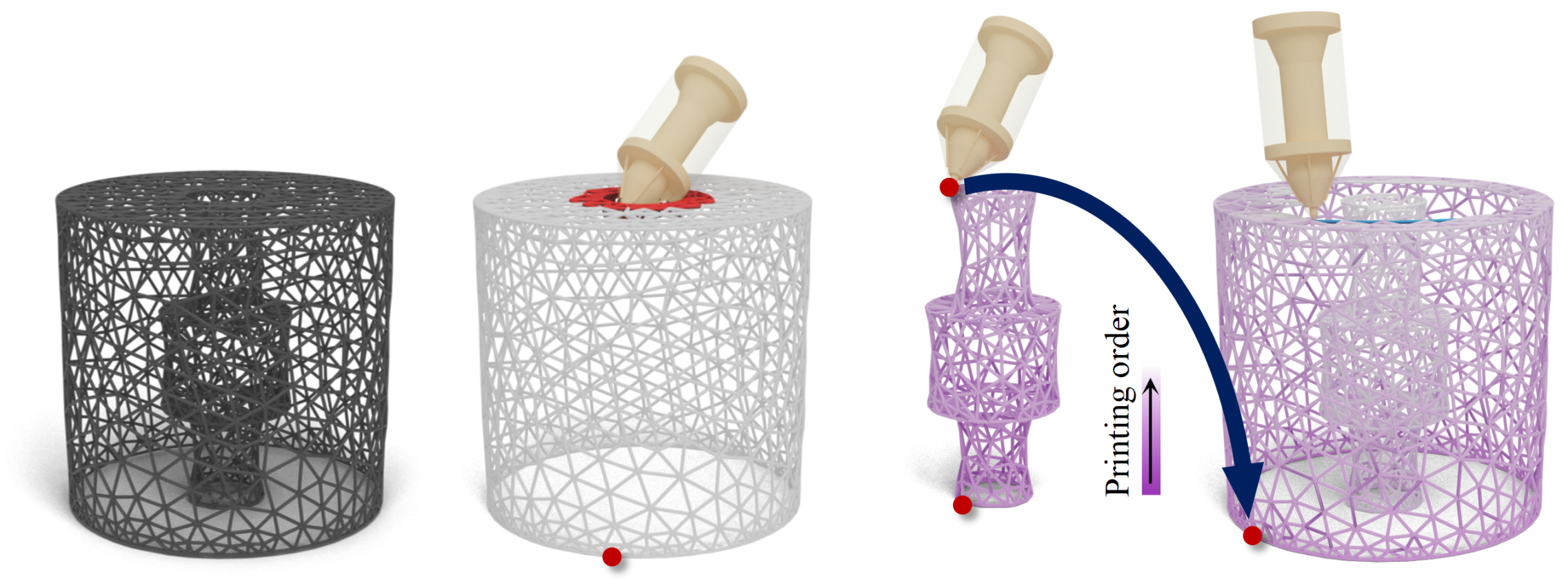}
\put(-242,4){\small \color{black}(a)}
\put(-180,4){\small \color{black}(b)}
\put(-115,4){\small \color{black}(c)}
\caption{For a given wireframe model as shown in (a) with its cavity extending into a through hole, selecting an appropriate starting node is important. As shown from (b), starting from a node on the outer structure can lead to a configuration with unavoidable collision. Differently, printing path for the model can be successfully planned when selecting a node from the inner structure as the starting node (c). Colors are employed to visualize the sequence of printing.}
\label{fig:startingNodeFeasibility}
\end{figure}

The starting node of planning is selected randomly by heuristics in our implementation (e.g., need to be a node on the ground for the wireframe printing). Before completing the process of path planning, the planner cannot predict if the randomly selected starting node can lead to a feasible solution (see Fig.\ref{fig:startingNodeFeasibility} for an example). Regarding this problem of feasibility, our algorithm incorporates an outer loop for offline computation applications. This allows initiating the algorithm from various random nodes, thereby generating multiple toolpaths from which the most suitable one is selected. If the algorithm starts from a node from the exterior of the model (Fig.\ref{fig:startingNodeFeasibility}(b)), once it yields a failure, a new starting node is selected. This process can be repeated until a feasible solution is found (e.g., the one as shown in Fig.\ref{fig:startingNodeFeasibility}(c)).

\subsection{Jumping action}
Besides CCF already discussed in Sec.~\ref{subsubsecCCF}, continuity is also preferred in wireframe and metal printing. Frequent jumps in wireframe printing will lead to stringing phenomena caused by the difficulty in controlling the extruder of a printer head (see the top row of Fig.\ref{fig:failureWireframePeeling}). For metal printing, melting powders at scattered points will lead to problems of low density and uneven surface finishes due to inconsistent melting and solidification of the material. For these reasons, we choose 'jumping' as a passive but not proactive action to be learned and controlled our algorithm.

Given the nature of the on-the-fly planner working on dynamically moved LSGs, our learning-based planner may drive the toolpath into topological obstacles -- i.e., no feasible next node can be found among the 1-ring neighbors of a center node $v_c$. We then need to apply some heuristic to `restart' the planner from a new node. 
\begin{itemize}
\item For wire-frame and CCF printing, we use the closest node of $v_c$ that has a few of its adjacent edges already included in $\mathcal{P}$. If no such node exist, we simply restart the planner on the uncovered part of the input graph.

\item For metallic printing, the farthest node on the graph that has not been included in $\mathcal{P}$ -- heuristically, the farthest point should have a very low temperature. 
\end{itemize}
Note that jumping only occurs when there is no possible next step, and it tends to move the center of the LSG outside the $n$-ring neighborhoods. This is considered a global decision, which is similar to picking a new starting node. The number of jumps is generally very small in experiments.

\begin{figure}
\centering
\includegraphics[width=\linewidth]{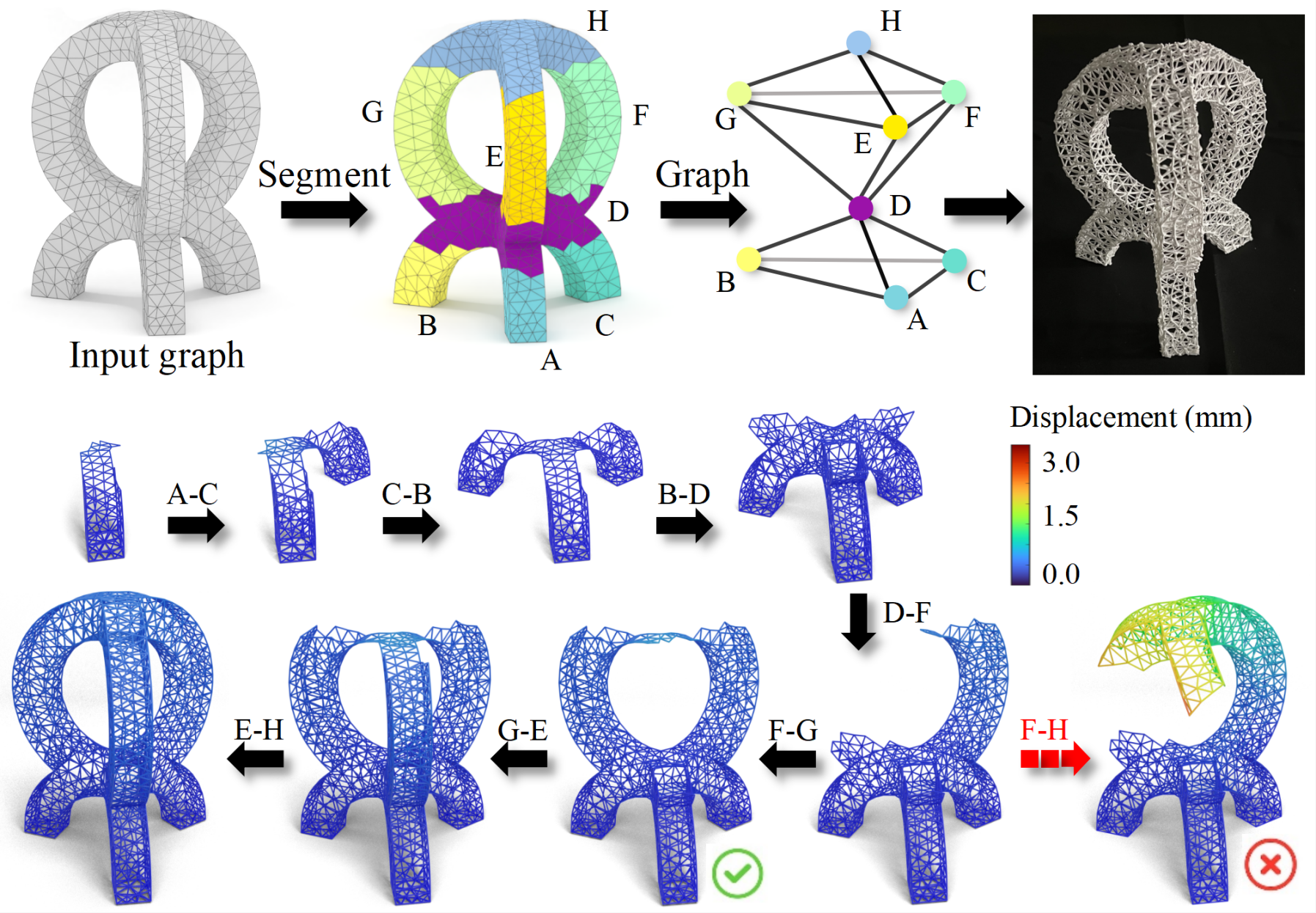}
\put(-242,101){\small \color{black}(a)}
\put(-50,101){\small \color{white}(c)}
\put(-242,3){\small \color{black}(b)}
\caption{An illustration of extension to capture global topological information by introducing graphs in multi-scales. (a) The Pavilion model is first segmented into different regions by topology analysis, and a graph at the coarse level can then be constructed by using each region as a node. (b) A brute-force search is then employed to determine the sequence of regions to print, where those steps with large deformation (e.g., F-H with displacement larger than $2.0 \mathrm{mm}$) will be prevented to avoid printing failure. (c) The final printing result of the Pavilion wire-frame. 
}\label{fig:MultiResPlanning}
\end{figure}

\subsection{Global topological information}
Our learning-based planner mainly considers the states inside an LSG although the historical intelligence can somewhat be learned and stored in the DQN repeatedly employed and updated. To overcome this limitation, a more global intelligence can be achieved by employing the graphs at different level of details. 

Taking the Pavilion model given in Fig.~\ref{fig:MultiResPlanning} as an example, we first conduct the Reeb-graph based topology analysis \cite{zhang2005feature} to segment the input model into 8 regions. A coarse-level graph can be constructed by these regions and their neighboring information. Each region becomes a node in the coarse-level graph, and an edge is constructed between two nodes if their corresponding regions share some common boundaries. We then apply the brute-force search to determine the best sequence giving the smallest deformation (as shown in Fig.~\ref{fig:MultiResPlanning}), where the collision detection is conducted by adding all the struts of a region as a whole. After figuring out the sequence with minimized deformation, our $Q$-learning based planner is applied to generate the toolpath in each region by the method introduced above. 

Note that the baseline of our DQN-based planner is the brute-force search. Our planner is designed to balance efficiency with performance, which does not outperform brute force search on smaller graphs (as already shown by the bar-charts in Figs.\ref{fig:Teaser} and \ref{fig:comparisonBFSCoralCat}). However, we can replace the brute-force search by our planner when there are large number of nodes on the coarse-level graph. 

\subsection{Global collision avoidance}
In the current implementation, the reward function of collision avoidance only considered the local collision between the swept volume of the printer head and the in-process structure. When using robotic hardware to realize the printing process, we heavily rely on the solver of \textit{inverse kinematics} (IK) to determine a solution that has no collision between the robotic arm and the in-process structures. Although we did not observe any problem when testing all the examples shown in this paper, a collision-free IK solution may not exist. A more sophisticated method needs to be developed in future research.

\section{Conclusion}\label{secConclusion}
In this paper, we proposed a learning-based planner that can work on large scale graphs with an on-the-fly method to construct the state space for the planner. A general framework for computing optimized 3D printing toolpaths has been developed. Our planner can cover different applications by defining their corresponding reward functions and state spaces, where the toolpath generation problems of wireframe printing, CCF printing, and metallic printing are selected to demonstrate its generality. The experimental results are quite encouraging in all these applications. Significant improvements can be observed.

\begin{acks}
    The project is supported by the chair professorship fund at the University of Manchester and UK Engineering and Physical Sciences Research Council (EPSRC) Fellowship Grant (Ref.\#: EP/X032213/1).
\end{acks}

\bibliographystyle{ACM-Reference-Format}
\bibliography{reference.bib}

\appendix
\section{Supplementary Document}
\subsection{Effectiveness of Pattern Encoding}
\begin{figure}
\centering
\includegraphics[width=\linewidth]{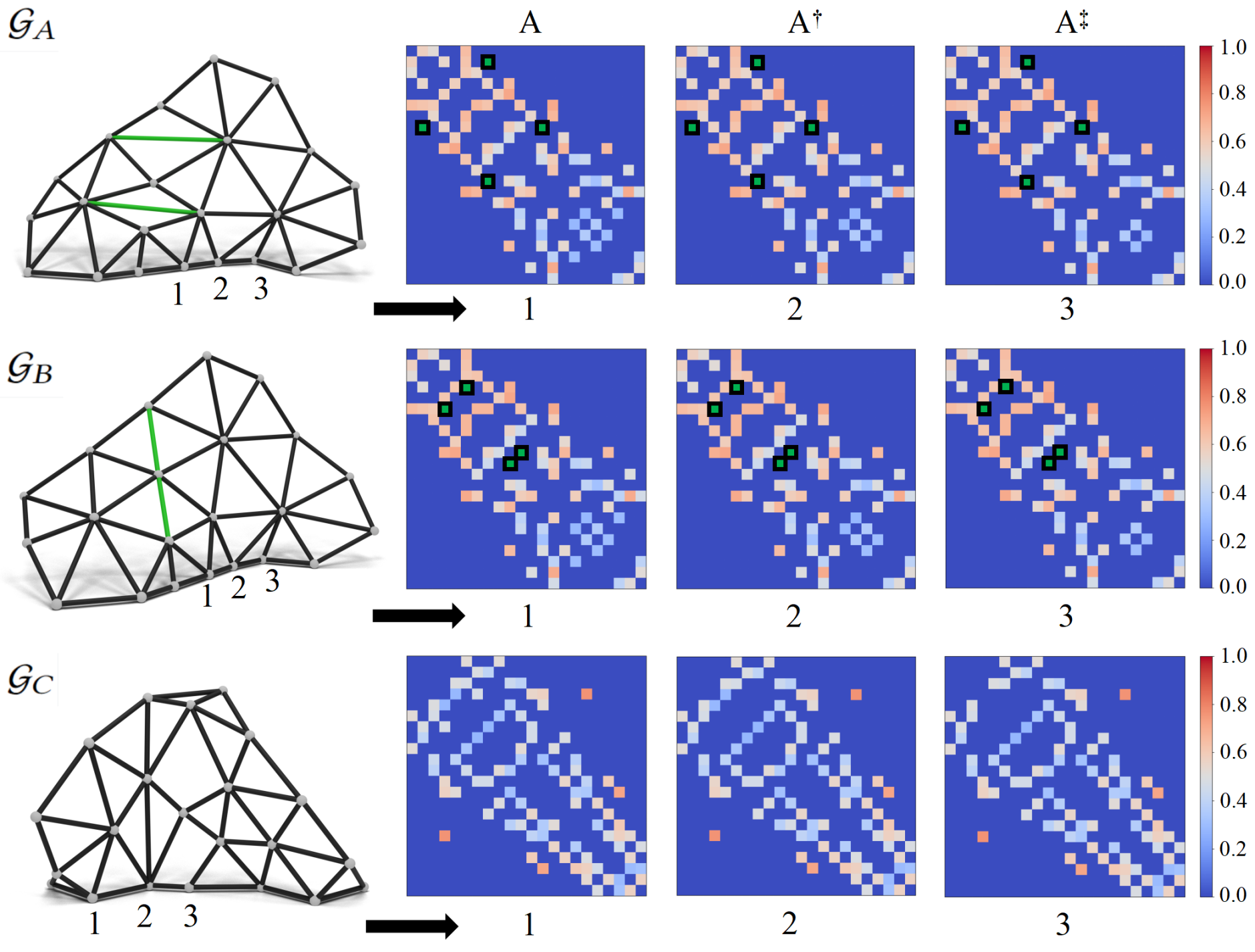}
\caption{
For two graphs $\mathcal{G}_A$ and $\mathcal{G}_B$ having slight different connectivities (i.e., those edges highlighted by green color), their corresponding adjacent matrices generated by our pattern encoding (P.E.) algorithm are similar to each other (i.e., only those highlighted green elements circled by black lines are different). When comparing with $\mathcal{G}_C$ that exhibits a markedly distinct topology, the adjacency matrices produced by our P.E. algorithm demonstrate remarkable divergence. The similarity evaluated by Eq.(1) between the states of $\mathcal{G}_A$ and $\mathcal{G}_B$ as $\mathbf{S}_A$ and $\mathbf{S}_B$ is $\rho(\mathbf{S}_A,\mathbf{S}_B)=0.95$ while $\rho(\mathbf{S}_A,\mathbf{S}_C)=0.42$ and $\rho(\mathbf{S}_B,\mathbf{S}_C)=0.44$.
}\label{fig:similar_subgraph}
\end{figure}

Studies have been conducted to verify the effectiveness of our pattern encoding (P.E.) algorithm. First of all, we demonstrate the similarity of adjacent matrices obtained from two similar graphs as $\mathcal{G}_A$ and $\mathcal{G}_B$ shown in Fig.~\ref{fig:similar_subgraph}, which have slightly different connectivities between nodes -- i.e., the edges highlighted by green color. Then, we conduct the comparison with a third graph $\mathcal{G}_C$ that has a markedly distinct topology, where the adjacent matrices generated by our algorithm are far more different. This has been verified by the quantitative measurement of similarity $\rho(\cdot,\cdot)$ evaluated by Eq.(1). 

\subsection{Stored networks as Prior}
\begin{figure}
\centering
\includegraphics[width=\linewidth]{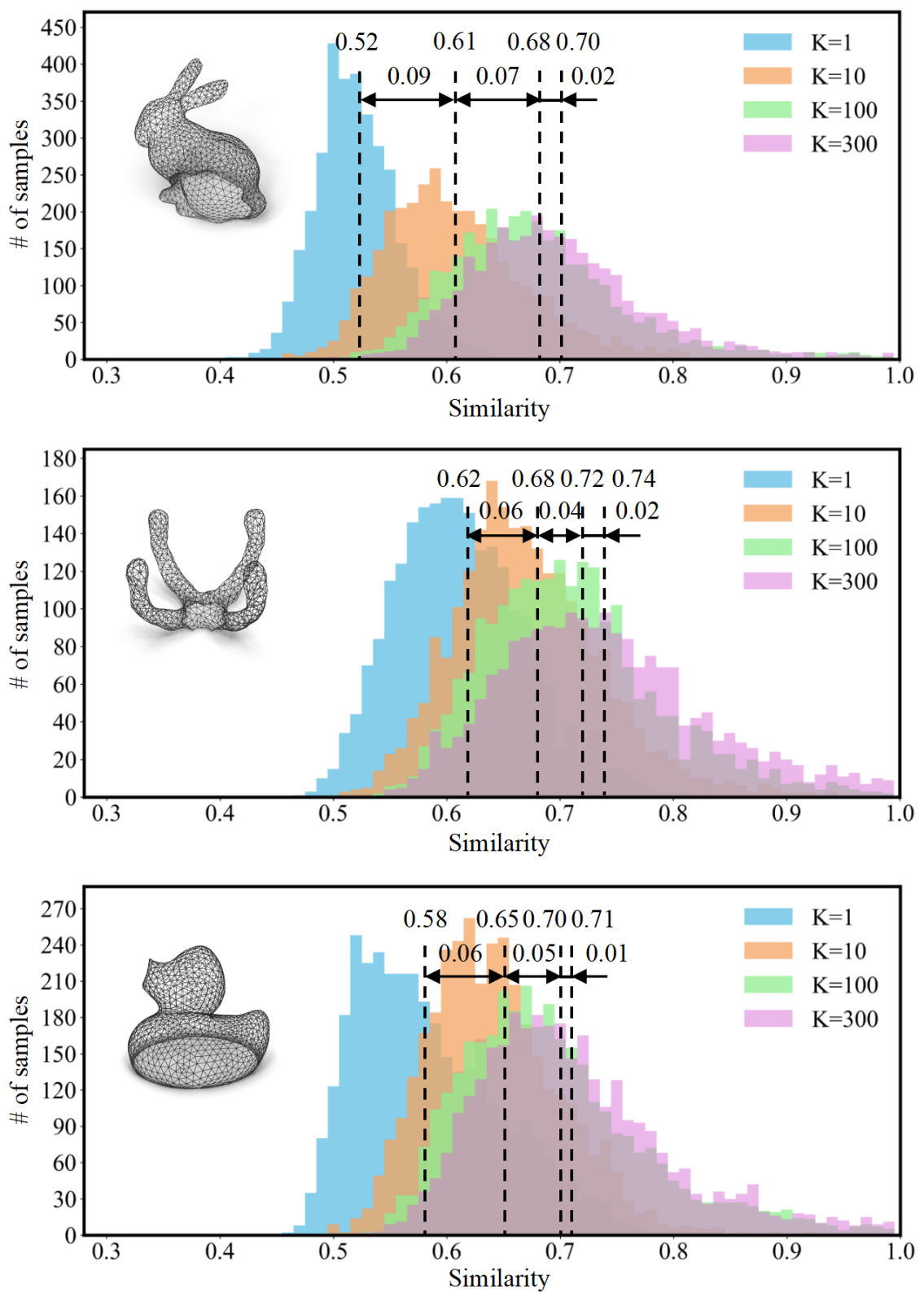}
\caption{
The histograms of $\max_{k=1,\ldots,K} \{ \rho_k \}$ when running the accelerated scheme with DQN prior reuse (Sec.~3.3) by using different value of $K$ (i.e., different numbers of stored DQNs as prior) on three models -- Bunny, Coral and Duck. The means of similarities are given by the dash lines. 
}\label{fig:pe number compare}
\end{figure}

Here we take some interesting tests to study the number of DQNs stored as prior to reuse in the accelerated scheme. Specifically, when generating a toolpath on a model, we collect the maximal similarity as $\max_{k=1,\ldots,K} \{ \rho_k \}$ on every LSGs and plot them as a histogram. The histograms of using different stored number of DQNs as priors are compared on three different models as shown in Fig.\ref{fig:pe number compare}. Taking the result of Bunny model as an example, the mean of similarities can be improved by around 17.3\% when increasing the value of $K$ from 1 to 10, and it can be further enlarged by another 11.5\% by increasing $K$ to 100. However, the improvement become less significant when using $K=300$. Similar trend can be observed on the other two models shown in Fig.\ref{fig:pe number compare}. As a result, $K=10$ is chosen according to these experimental tests to balance the effectiveness and the cost of memory. 

\begin{figure}
\centering
\includegraphics[width=\linewidth]{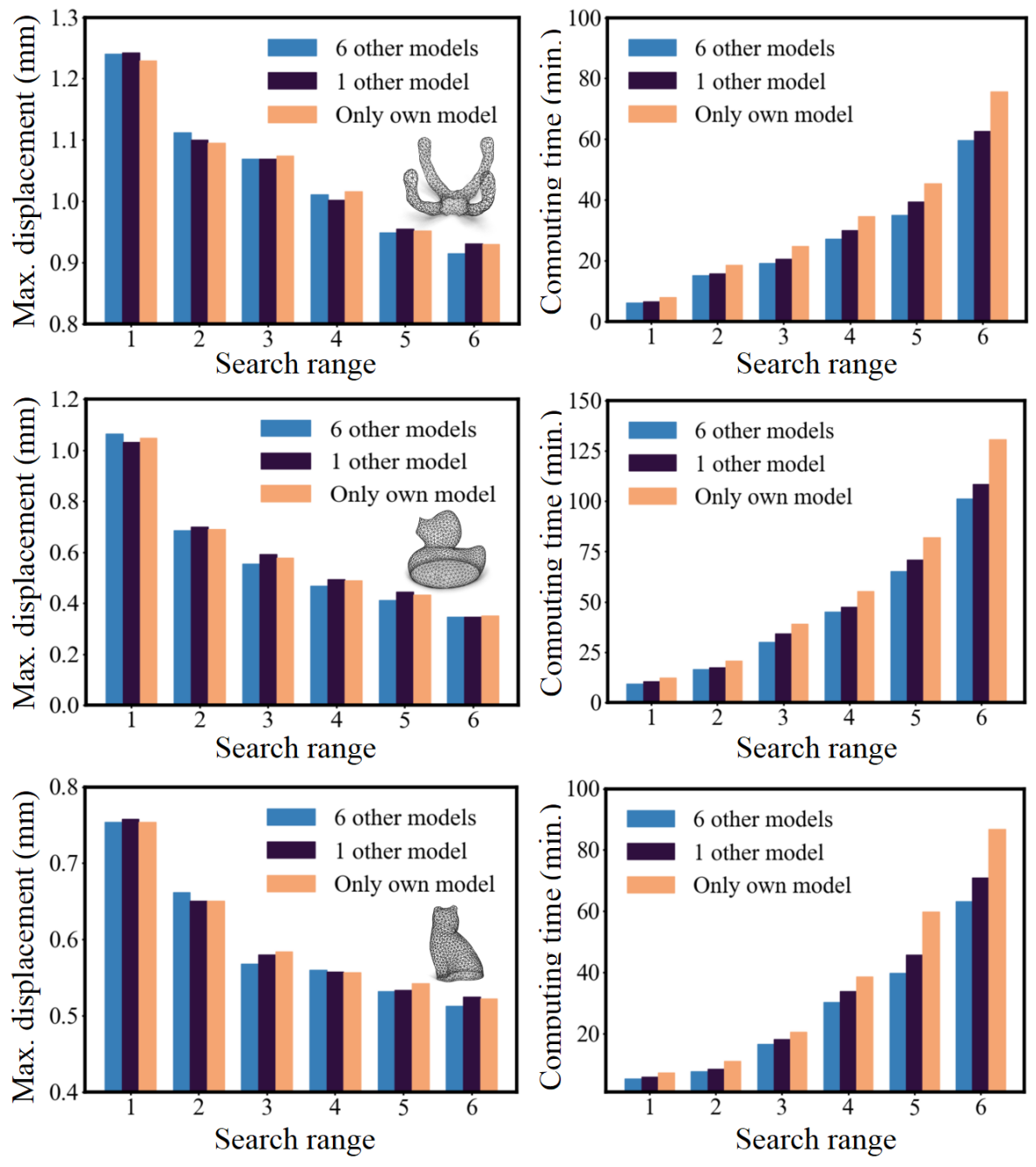}
\caption{
The learning process can be effectively accelerated when more other models have been learned to generate the initial DQNs stored as prior. Different colors indicate the prior is obtained from different number of other models that the planner has `seen'.
}\label{fig:initial prior compare}
\end{figure}

The other interesting study is whether the learning process on a new model can benefit from the priors learned on other models. The statistical results are collected on three different models (Coral, Duck and Cat) by using the priors learned from i) the own model, ii) 1 other model (Bunny) and iii) 6 other models (Femur, Coral, Bunny, Duck, Cat and Molar). The planner is studied by using different search ranges $n = 1, \cdots, 6$. As can be observed from Fig.\ref{fig:initial prior compare}, the learning process can be further accelerated when the priors have been learned from more other models. Note that in these tests, $K=10$ is employed. On the other aspect, our $Q$-learning based planner can handle diverse graphs well even for the tests taken by using the priors obtained from the earlier LSGs on the own model (see the results in Fig.\ref{fig:initial prior compare} labelled as `Only own model' and also other results shown later in Figs.\ref{fig:10 sets of data} \& \ref{fig:10 sets of data ccf} when comparing with the BFS-based results).

\begin{figure}
\centering
\includegraphics[width=\linewidth]{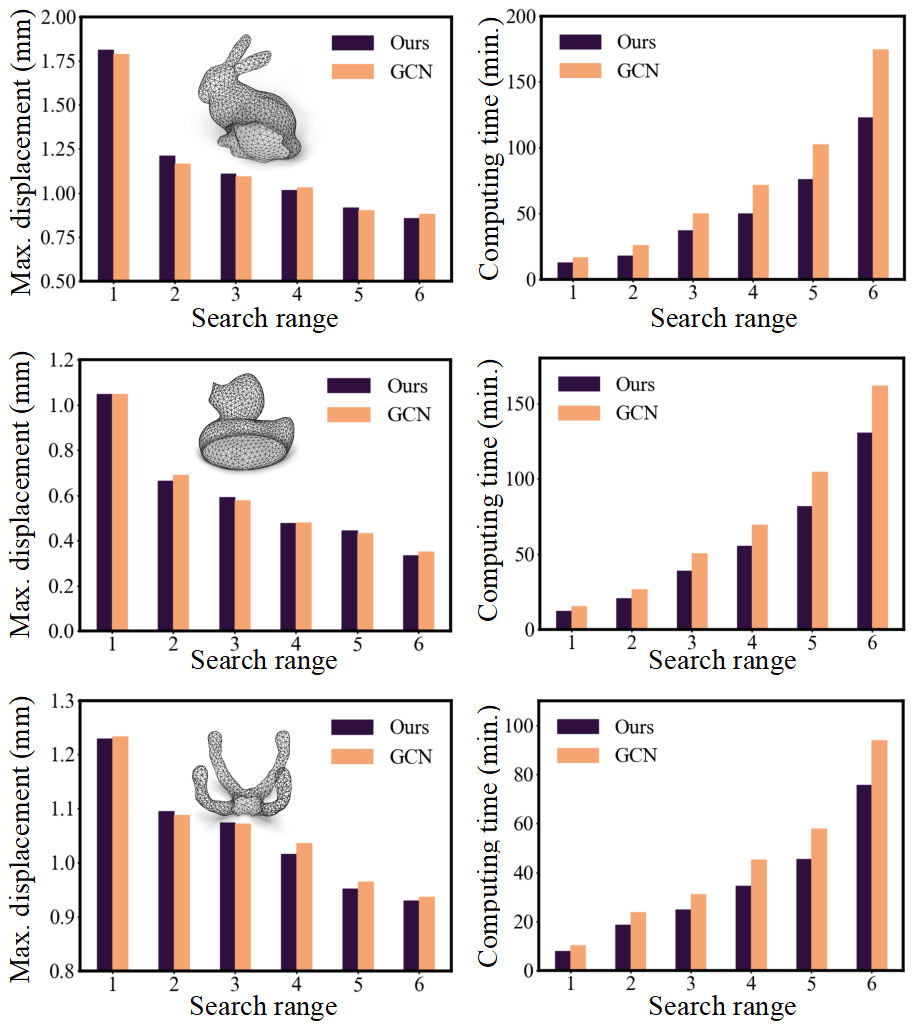}
\caption{Comparison of GCN-based and our CNN-based implementation for the $Q$-learning based planner in terms of performance and computation time on different models. It can be found that both give similar results but ours is 19.4\% to 29.6\% faster. 
}\label{fig:GCN compare}
\end{figure}

\subsection{GCN VS CNN}
We compare the our implementation of CNN-based DQNs with the Graph Convolution Network (GCN) based DQNs using the common benchmark library PyG \cite{Fey/Lenssen/2019}. Note that the row-column convolution operators are employed in our CNN-based DQNs. Three models have been tested as shown in Fig.\ref{fig:GCN compare}. We find that GCN-based and CNN-based DQNs can achieve similar performance -- in many cases, our CNN-based DQNs give slightly better results; however, in terms of computation time, our CNN-based implementation is 19.4\% to 29.6\% faster than GCN due to the more efficient memory visit in the matrix-based implementation

\subsection{Study of Network structure}
In the current implementation for the DQNs of our learning-based planner, two \textit{Edge-to-Edge} (E2E) convolution layers and one \textit{Edge-to-Node} (E2N) convolution layers are employed. Both are based on the row-column convolution operators introduced in \cite{KAWAHARA20171038}. We conducted a few experiments to decide the number of layers. 

\begin{figure}
\centering
\includegraphics[width=\linewidth]{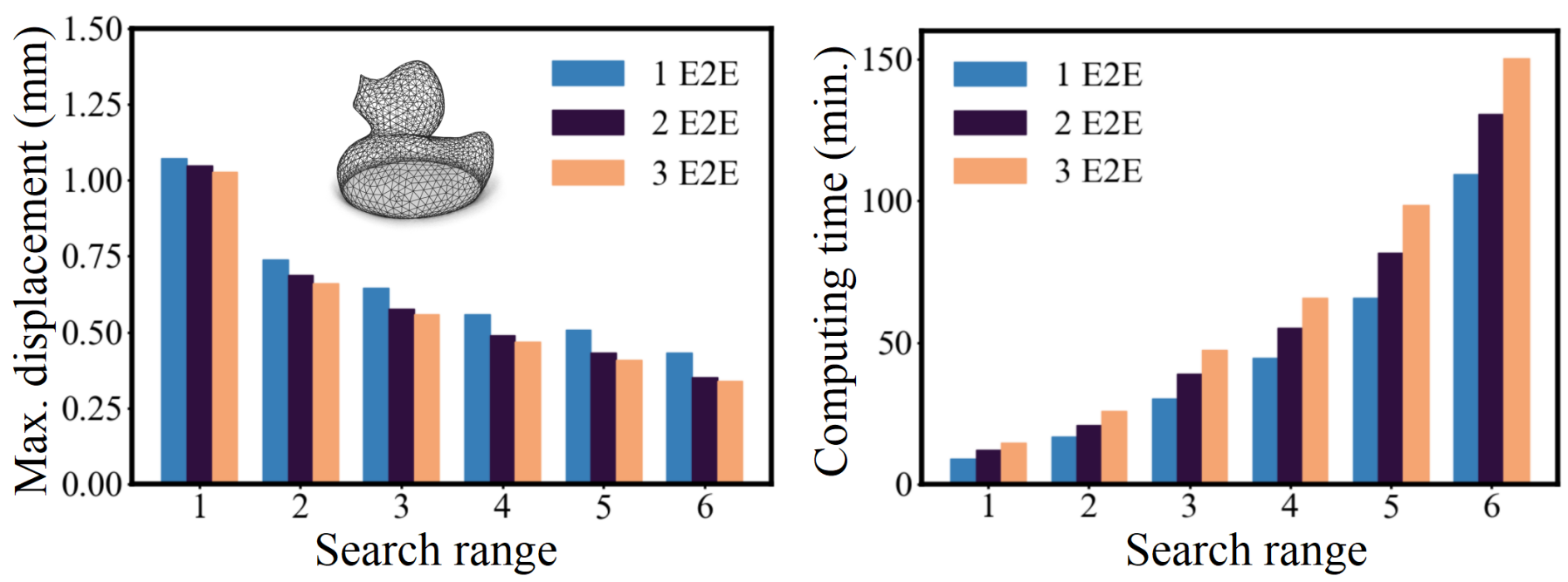}
\caption{Comparison in performance and computational time of using different number of E2E layers.
}\label{fig:network layer compare}
\end{figure}

The first experiment is to test the performance (i.e., the maximal displacement for the wire-frame printing) and the computation time when using different number of E2E layers. The results are given in Fig.\ref{fig:network layer compare}. It can be found that using three E2E layers can slightly improve the performance but will consume much more computing time. According this experiment, two E2E layers are selected in our final implementation.

\begin{figure}
\centering
\includegraphics[width=\linewidth]{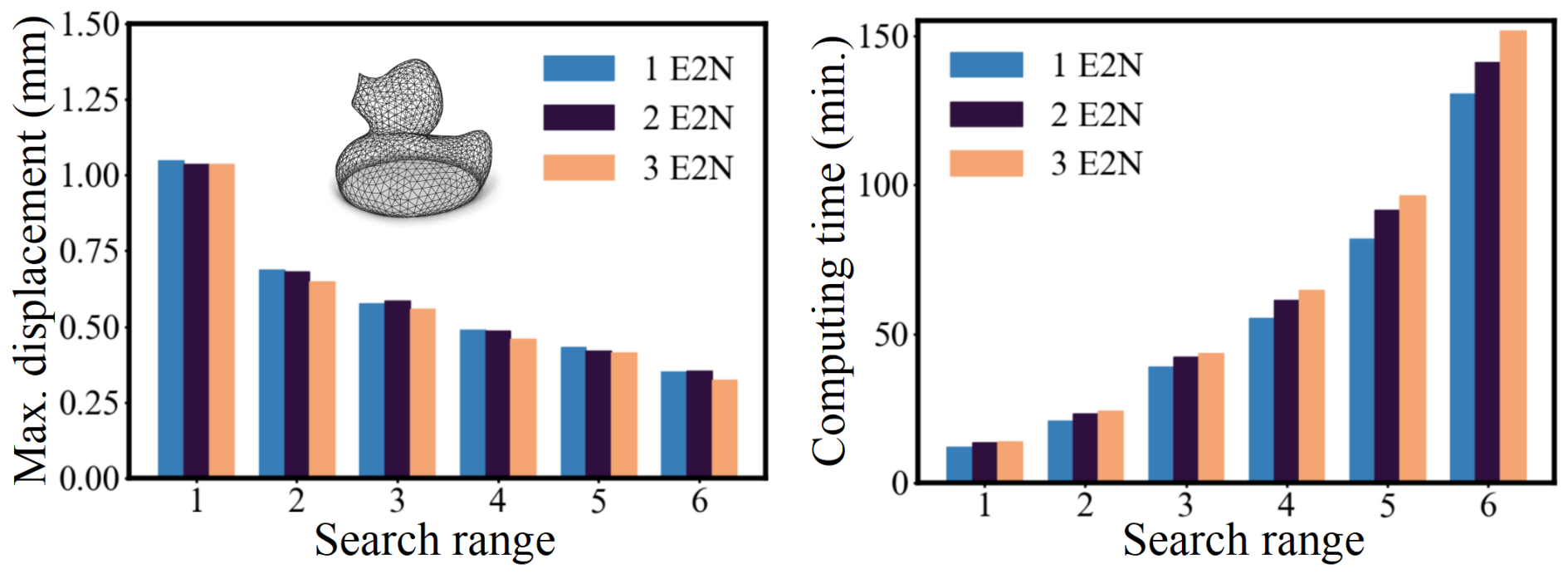}
\caption{Comparison in performance and computational time of using different number of E2N layers.
}\label{fig:network layer compare2}
\end{figure}

The other experiment is for evaluting if the performance can be further improved by using more E2N layers. Again, both the performance and the computational time are studied. The results are given in Fig.\ref{fig:network layer compare2}. It is found that adding more E2N layers brings insignificant improvement but takes more computational time. Therefore, we adopt only one E2N layer in our Q-learning based path planner.

\subsection{Effectiveness of historical information}
\begin{figure}
\centering
\includegraphics[width=\linewidth]{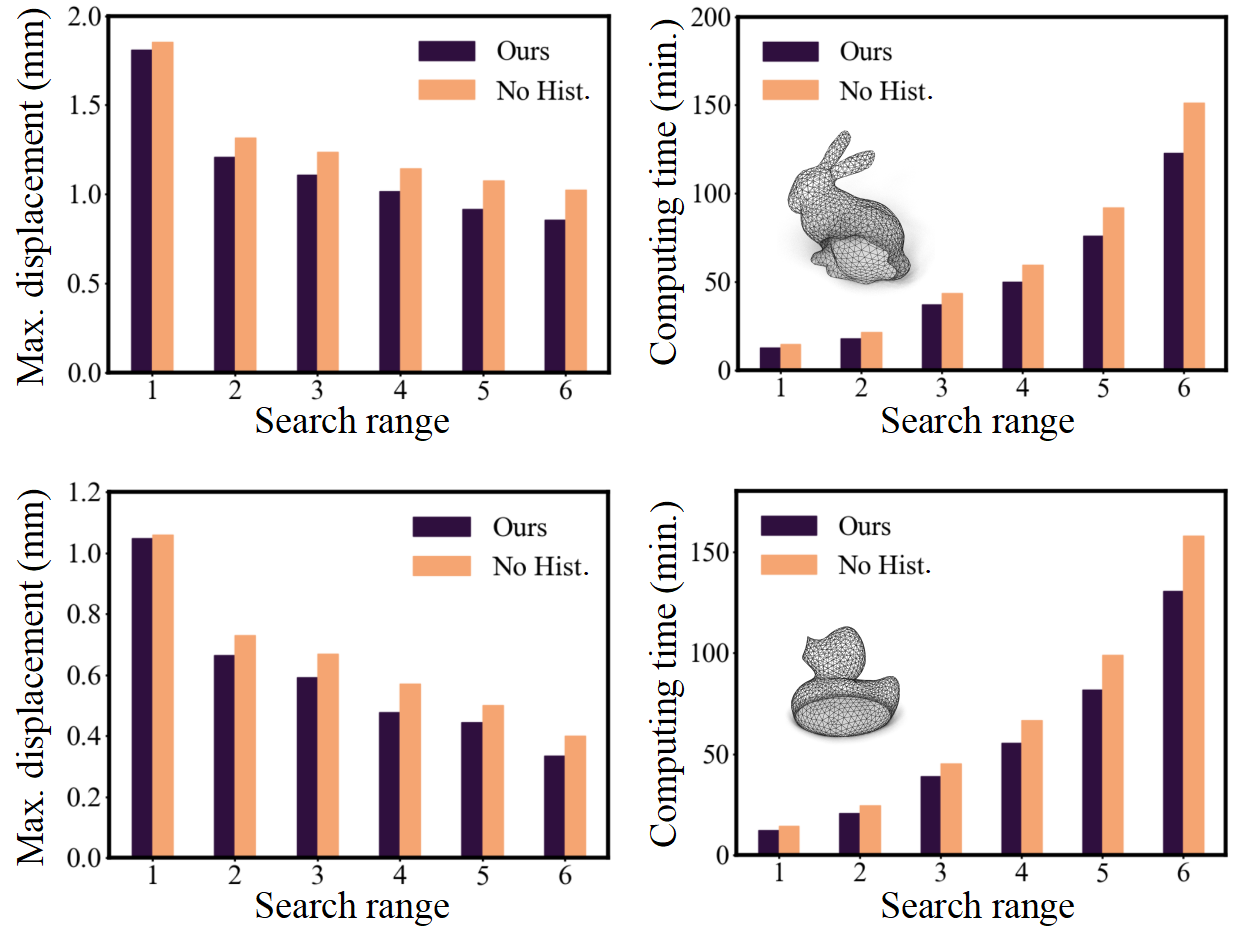}
\caption{\rev{The performance and the computing time in wire-frame printing are compared by using our enriched states v.s. the simple state without historical information (denoted by `No Hist.').}}
\label{fig:history size}
\end{figure}

\rev{In our proposed approach, short-term memory information has been introduced to enrich the features learned by DQNs for better capturing the dependencies between multiple states and the dynamic changes of the printing process. Specifically, we employ an enriched state as $\mathbf{S}=[\mathbf{A}, \mathbf{A}^{\dag}, \mathbf{A}^{\ddag}]$ (see Sec.~3.3.2). We have conducted tests on the Bunny model and the Duck model to demonstrate the advantage of using this state with short-term memory information vs. the state $\mathbf{S}=\mathbf{A}$ without history information (denoted as `No Hist.'). The results in terms of deformation have been shown in the left of Fig.\ref{fig:history size} by using different search ranges for LSGs. Better results with smaller displacements are obtained by using our enriched state.}

\rev{Moreover, the enriched states can provide more stable gradient updates in the early stage of DQN-based learning. That means the interior learning routine converges faster and therefore can compute the resultant toolpath faster. The statistics of computing times have been given in the right of Fig.\ref{fig:history size} -- again tested by using LSGs with different search ranges.}

\begin{figure}
\centering
\includegraphics[width=\linewidth]{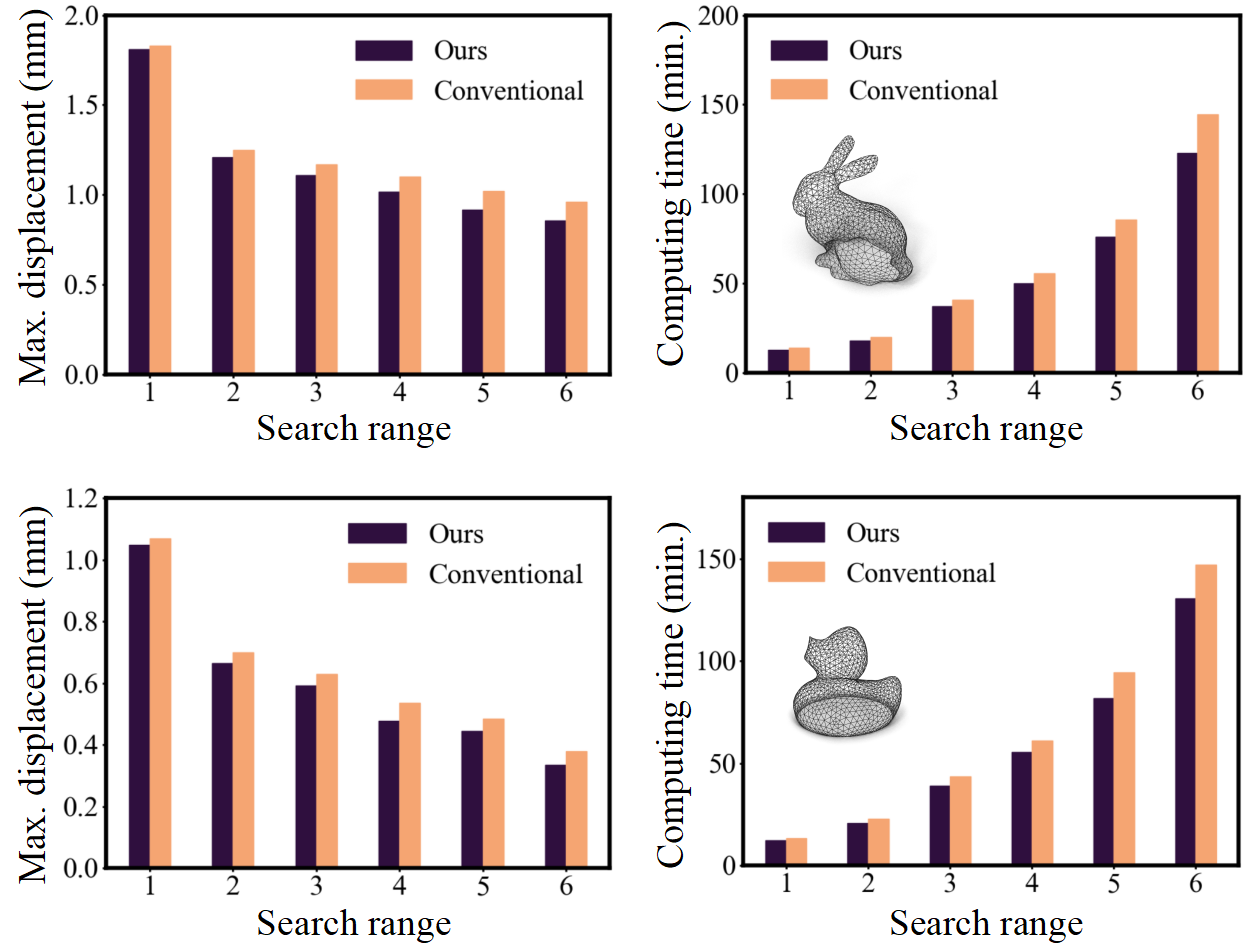}
\caption{\rev{Comparison of the performance and the computing time by using the exponential discount function (denoted by `Conventional') v.s. the Gaussian-like discount function used in our implementation.}}
\label{fig:discount function}
\end{figure}

\begin{figure}
\centering
\includegraphics[width=\linewidth]{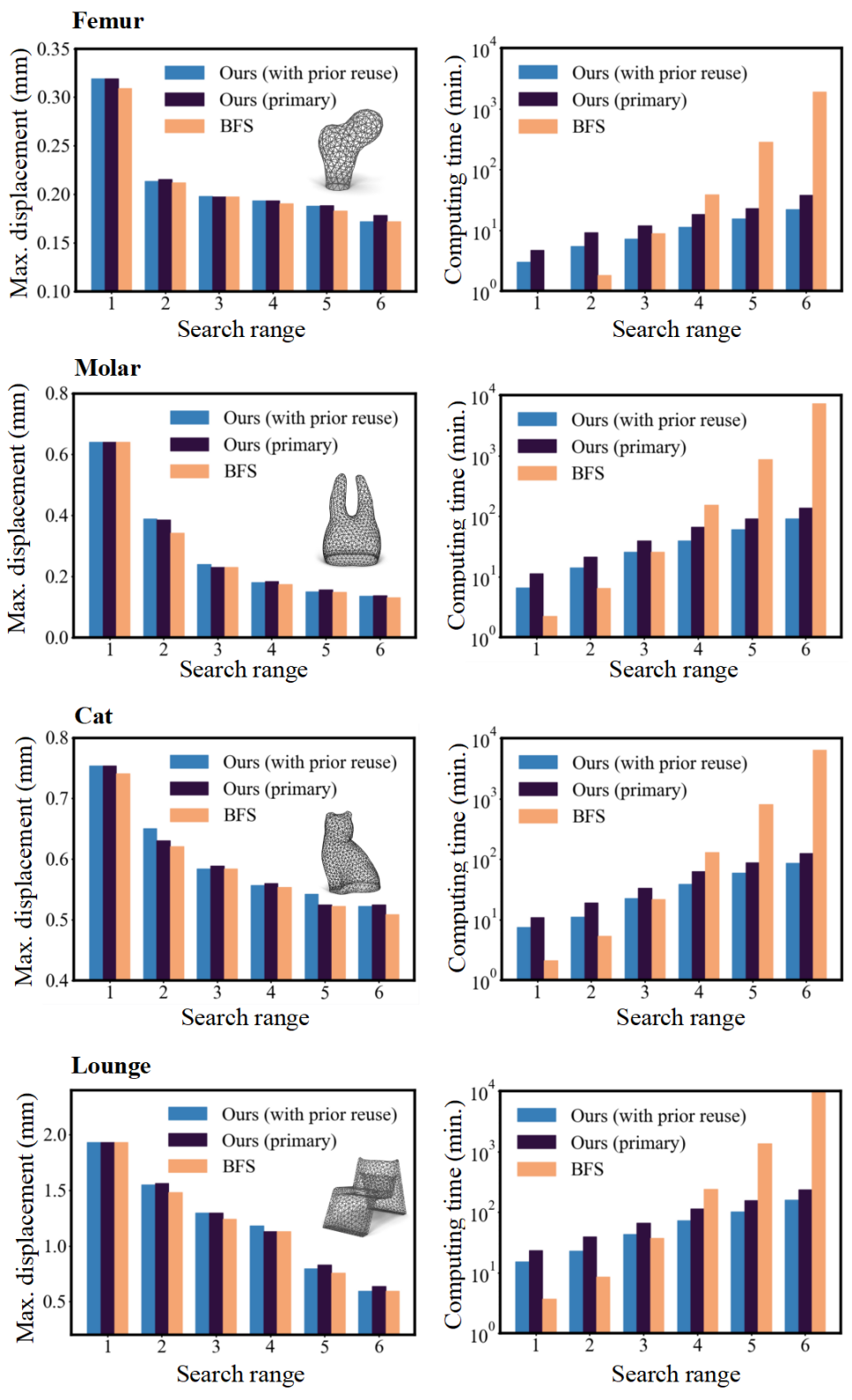}
\caption{Comparison of the performance and the computing time in wireframe printing by using our learning based planner with and without prior reuse (primary scheme), and also the BFS method. Please note that the time axis is represented on a logarithmic scale.}\label{fig:10 sets of data}
\end{figure}

\subsection{Study of discount factor formulation}
\rev{In conventional DQN-based reinforcement learning (e.g., \cite{silver2016mastering,silver2017mastering}), the discount function as follows is employed.}
\begin{equation}\label{eq:DecayFunc2} 
    \Gamma(i)= 0.9 ^{i}
\end{equation}
\rev{which generates an exponentially decreasing impact of rewards on the $i$-th step ($i=1,2,\cdots$). When using this exponential discount function, the learned model will have more focus on distance future states. Differently, our implementation is based on a Gaussian-like discount function as given in Eq.(4) of Sec.~4.2. This gives a large reward weight in the early stage so that the model can effectively learn the importance of the decision in the early stage. For our LSG-based on-the-fly planner, this change can accelerates the computation by up to 15.01\%. Moreover, the objective of DQN-based learning in our problem is to find the `best' next step, our discount function gives priority to the earlier rewards. The performance is also improved by up to 11.35\% as shown in Fig.\ref{fig:discount function} when using the Gaussian-like discount function.}

\begin{figure}
\centering
\includegraphics[width=\linewidth]{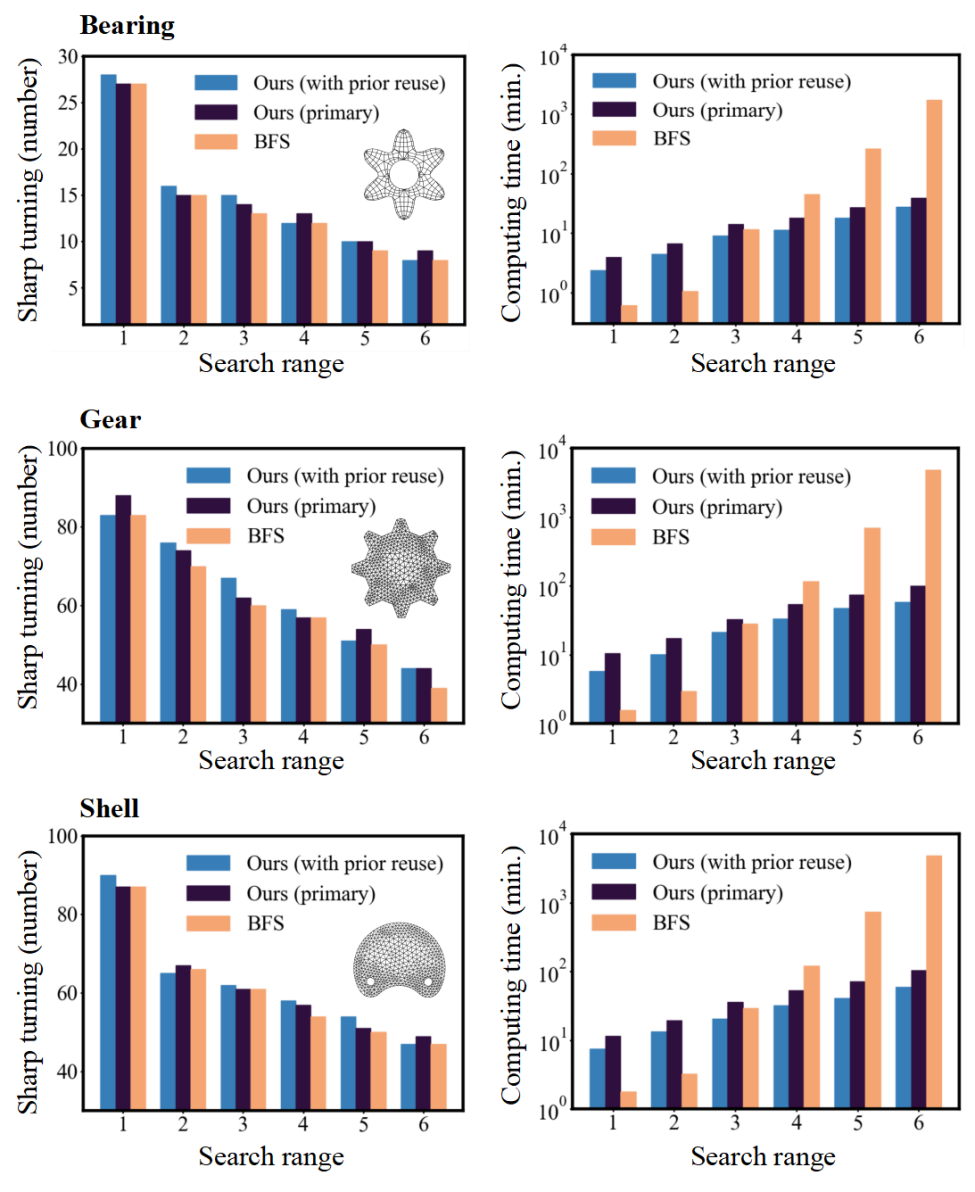}
\caption{The performance and the computing time in CCF printing are compared by using our learning based planner with and without prior reuse (primary scheme), and also the BFS method. Again the time axis is represented on a logarithmic scale.}
\label{fig:10 sets of data ccf}
\end{figure}

\subsection{Results on more examples}
We have tested our $Q$-learning based planner on a variety models for the wireframe printing and the CCF printing. The results have been compared with those obtained from Brute-Force Search (BFS) based method in terms of both the performance and the computing time. Both the results with and without the prior reuse are given in these results -- see Fig.\ref{fig:10 sets of data} for the detailed comparison for wireframe printing and Fig.\ref{fig:10 sets of data ccf} for the comparion for CCF printing. Our algorithm (regardless of the existence of prior reuse) needs significantly less computing time while providing similar performance as the BFS based method.

\clearpage

\end{document}